\documentclass{article}
\PassOptionsToPackage{numbers, compress}{natbib}

\usepackage[final]{neurips_2025}

\usepackage[utf8]{inputenc} 
\usepackage[T1]{fontenc}    
\usepackage{url}            
\usepackage{booktabs}       
\usepackage{amsfonts}       
\usepackage{nicefrac}       
\usepackage{microtype}      
\usepackage{xcolor}         

\usepackage{bm}
\usepackage{array}
\usepackage{multirow}
\usepackage{threeparttable}

\usepackage{amsmath}
\allowdisplaybreaks
\usepackage{amssymb}
\usepackage{mathtools}
\usepackage{amsthm}

\usepackage{microtype}
\usepackage{graphicx}
\usepackage{subfig}
\usepackage{booktabs} 

\usepackage{wrapfig}
\usepackage[hidelinks]{hyperref} 

\hypersetup{
    colorlinks=true,  
    linkcolor=purple,  
    citecolor=blue,   
    urlcolor=blue,    
    filecolor=blue,   
    pdfborder={0 0 0} 
}

\usepackage[capitalize,noabbrev]{cleveref}

\theoremstyle{plain}
\newtheorem{theorem}{Theorem}[section]
\newtheorem{proposition}[theorem]{Proposition}
\newtheorem{lemma}[theorem]{Lemma}

\newtheorem{definition}[theorem]{Definition}
\theoremstyle{definition}
\newtheorem{assumption}{Assumption}[section]
\theoremstyle{definition}
\newtheorem{remark}{Remark}[section]

\usepackage[textsize=tiny]{todonotes}

\usepackage{enumitem}

\title{Generalization Error Analysis for Selective State-Space Models Through the Lens of Attention}

\author{
  Arya Honarpisheh
  \qquad
  Mustafa Bozdag
  \qquad
  Octavia Camps
  \qquad
  Mario Sznaier \vspace{3pt} \\ 
  Department of Electrical and Computer Engineering \\
  Northeastern University, Boston, MA 02115 \vspace{3pt} \\
  \texttt{\{honarpisheh.a, bozdag.m\}@northeastern.edu},\\ 
  \texttt{\{camps, msznaier\}@coe.neu.edu} 
}

\begin{document}

\maketitle

\begin{abstract}
       State-space models (SSMs) have recently emerged as a compelling alternative to Transformers for sequence modeling tasks. This paper presents a theoretical generalization analysis of selective SSMs, the core architectural component behind the Mamba model. We derive a novel covering number-based generalization bound for selective SSMs, building upon recent theoretical advances in the analysis of Transformer models. Using this result, we analyze how the spectral abscissa of the continuous-time state matrix influences the model’s stability during training and its ability to generalize across sequence lengths. We empirically validate our findings on a synthetic majority task, the IMDb sentiment classification benchmark, and the ListOps task, demonstrating how our theoretical insights translate into practical model behavior.
\end{abstract}


\section{Introduction}
\label{sec:intro}
Foundation models trained on large-scale datasets have become the dominant paradigm in modern machine learning following the introduction of the Transformer architecture \cite{vaswani2017attention}. Although they are very effective in sequence processing, the global nature of self-attention imposes limitations: a finite sequence window and quadratic scaling w.r.t. the window length. Recently, a new family of models named state-space models (SSMs) became a popular alternative to address this problem \cite{gu2021ssm, gu2022s4}. These models are rooted in the classical state-space representations from control theory introduced by \citet{kalman1960new}. Theoretical analysis of SSMs has primarily focused on linear time-invariant (LTI) settings. \citet{racz2024length} derive a length-independent generalization bound for deep stable LTI SSMs using Rademacher contraction techniques, while \citet{liu2024generalization} establishes a length-dependent bound and proposes corresponding initialization and regularization schemes. These results rely on system stability from control theory, ensuring that operator norms remain well-defined and finite. LTI architectures allow for an easy imposition of the stability assumption, making them well-suited for bounded-norm analysis.

The state-of-the-art SSMs that form the core of the Mamba and Mamba-2 architectures are selective and inherently nonlinear \cite{gu2024mamba, dao2024transformers_are_ssms}. They employ a selective-scan mechanism to capture long-term dependencies, allowing for adaptive and efficient sequence processing. This design, however, makes their analysis distinct from LTI SSMs and more closely aligned with self-attention. Despite their growing empirical success, theoretical understanding of their capabilities remains limited, with the exceptions of \citet{ali2024hidden} and \citet{dao2024transformers_are_ssms}, who establish a connection between Transformers and SSMs, \citet{anonymous2025hope} who studies the initialization and training behavior of SSMs, and other recent works that examine their expressive power \cite{cirone2024theoretical_selective_ssm, sarrof2024expressive, nandakumar2025expressive_gram, terzic2025expressiveness, yu2025block}. 

Alongside expressive power, which characterizes model bias, an equally important question is that of generalization, which captures the variance component of the bias–variance tradeoff, and it is central in theoretical machine learning. Recently, generalization bounds have been developed across various domains, including graph neural networks \citep{wang2025gen_gnn}, large language models \citep{lotfi2024gen_llm}, meta‐learning \citep{wang2024gen_meta}, recommender systems \citep{zhang2024gen_recommender}, representation learning \citep{zhang2024gen_representaion}, and various neural network architectures \citep{andreeva2024gen_topological, zhang2024gen_shallow_NN, tadipatri2025gen_convex}. Work on the generalization capabilities of attention models often builds on the covering‐based framework of \citet{zhang2002covering} and \citet{bartlett2017spectrally}, originally developed for linear models and deep networks. Recent adaptations include \citet{edelman2022inductive}, which studies inductive bias in Transformers; \citet{trauger2024length_independent_transformer}, which establishes length‐independent bounds; and \citet{truong2024rank}, which derives rank‐dependent bounds.

In this work, we investigate the generalization properties of selective SSMs. Unlike prior work on LTI SSMs that relies on classical tools from control theory and stability assumptions, our approach builds on recent theoretical developments for Transformers and self-attention mechanisms. By unrolling the selective SSM into an attention-like formulation, we enable the use of covering number techniques originally developed for Transformer models. This connection introduces new technical challenges: selective SSMs feature input-conditioned dynamics, input-projected $\bm{B}$ and $\bm{C}$ matrices, and discretization from continuous-time systems. We address these by developing a unified covering argument that combines tools from both RNN and attention-based analyses. In particular, we handle the state matrix $\bm{A}$ through stability and discretization, while treating the input projections $W_{\bm{B}}$ and $W_{\bm{C}}$ as linear function classes, analogous to the key-query structure in attention. This results in a two-tiered covering construction that captures the hybrid structure of selective SSMs and is central to our generalization bound, allowing us to study how model characteristics influence generalization and training behavior. We support our theoretical findings with experiments on synthetic and real-world sequence classification tasks. Our main contributions can be summarized as follows:  

\begin{itemize}[leftmargin=4pt, itemsep=2pt, topsep=0pt]
\item[] \textbf{Theoretically}, we derive a novel covering number-based generalization bound for selective SSMs by unrolling their structure and leveraging their connection to attention mechanisms. We also provide a bound for linear attention to rigorously demonstrate the link between the two model classes.
\item[] \textbf{Analytically}, we investigate the implications of our bound to understand how generalization depends on the sequence length $T$ and the spectral abscissa $s_{\bm{A}}$ of the state matrix. In particular, we show how training can fail to stabilize unstable models when $T$ is large, resulting in poor generalization despite the expressivity of the model.
\item[] \textbf{Empirically}, we validate our analysis on a synthetic majority task, the IMDb sentiment classification benchmark, and the ListOps task. These experiments demonstrate how model behavior varies with sequence length and stability, highlighting the practical impact of our theoretical insights.
\end{itemize}


\section{Preliminaries}
\label{sec:prelim}

\subsection{Notation}
\label{sec:notation}

We denote the real and imaginary parts of a complex number by $\Re(\cdot)$ and $\Im(\cdot)$. We use $\mathbb{R}, \mathbb{R}^n, \mathbb{R}^{n \times m}$ to denote the set of real numbers, $n$-dimensional real vectors, and $n \times m$ real-valued matrices, respectively. Lowercase letters like $x$ refer to scalars or vectors, while bold uppercase letters like $\bm{X}$ denote matrices. The $\ell_p$-norm of a vector $x$ is written as $\|x\|_p$, and the $p$-operator norm of a matrix $\bm{X}$ is written as $\|\bm{X}\|_p$. For a matrix $\bm{X}$ with column vectors $x_1, \ldots, x_n$, the mixed norm $\|\bm{X}\|_{p,q}$ is defined as $\left\|\,\begin{bmatrix} \|x_1\|_p & \cdots & \|x_n\|_p \end{bmatrix}\,\right\|_q$. The Kronecker product of matrices $\bm{X}$ and $\bm{Y}$ is denoted by $\bm{X} \otimes \bm{Y}$. Continuous-time state-space matrices are written as $\bm{A}_c, \bm{B}_c, \bm{C}_c$, while their discrete-time counterparts are denoted by $\bm{A}, \bm{B}, \bm{C}$. The variable $N$ represents the number of states per channel, $T$ is the sequence length, $d$ is the number of channels, and $m$ is the number of samples. We use $(t)$ to denote continuous time indexing and $[t]$ for discrete time. Lastly, we denote by $ s_{\bm{A}} := \max_i \{ \Re(\lambda_i(\bm{A})) \}$ the \emph{spectral abscissa} of a matrix  $\bm{A} \in \mathbb{R}^{n \times n}$, the largest real part among the eigenvalues of $\bm{A}$. This quantity is commonly used in stability analysis, since a continuous-time system with state matrix $\bm{A}$ is asymptotically stable in the sense of Lyapunov if and only if $s_{\bm{A}} < 0$.

\subsection{State-Space Models}
\label{sec:ssm}
LTI SSMs are based on the classical single-input single-output (SISO) state-space representation:
\begin{equation}
    \begin{aligned}
        \dot{x}^{(j)}(t) &= \bm{A}_c^{(j)} x^{(j)}(t) + \bm{B}_c^{(j)} u_j(t) \\
        y_j(t) &= \bm{C}_c^{(j)} x^{(j)}(t)
    \end{aligned}
    \label{eq:state_space}
\end{equation}

where $\bm{A}_c^{(j)}\in\mathbb{R}^{N \times N}$, $\bm{B}_c^{(j)} \in\mathbb{R}^{N \times 1}$, $\bm{C}_c^{(j)} \in\mathbb{R}^{1 \times N}$, and $D_c^{(j)} \in\mathbb{R}$ represent the dynamics of a continuous-time LTI system corresponding to the \(j\)\textsuperscript{th} channel (feature) $u_j(t)$ of the input signal $u(t)$ using the hidden states $x^{(j)}(t)\in\mathbb{R}^N$.
With the notable exception of S5 \cite{smith2023s5}, most SSM implementations use $d$ SISO SSMs as in \eqref{eq:state_space} in a single block, one for each channel respectively.

\textbf{Selective SSMs}, introduced with Mamba \cite{gu2024mamba}, make the model parameters and the discretization step size input-dependent to increase expressive power. In particular, for the \(j\)\textsuperscript{th} channel, the continuous-time state-space parameters \(\bm{B}_c^{(j)}(t)\), \(\bm{C}_c^{(j)}(t)\), and the step size \(\Delta(t)\) are:
\begin{equation}
\label{eq:Mamba_projections}
\begin{aligned}
    \begin{array}{l}
        \bm{B}_c^{(j)}(t) = \bm{W}_B\,u(t) \\
        \bm{C}_c^{(j)}(t) = u(t)^\top \bm{W}_C^\top
    \end{array}
    \quad
    \vcenter{
        \hbox{
            $\Delta(t) = \tau_\Delta\bigl(p + q^\top u(t)\bigr)$
        }
    }
\end{aligned}
\end{equation}
Here,  $\bm{W_B},\bm{W_C}\in\mathbb{R}^{N\times d}, q \in \mathbb{R}^{d}$ and $p \in \mathbb{R}$ are learnable weights and $\tau_\Delta(x) = \ln(1 + e^x)$ is the softplus function. The input and output matrices $\bm{B}_c^{(j)}$ and $\bm{C}_c^{(j)}$ are linear projections of the input $u(t)$. 
Thus, the state-space model for the \(j\)\textsuperscript{th} channel is:
\begin{equation}
\begin{aligned}
    \dot{x}^{(j)}(t) &= \bm{A}_c^{(j)} x^{(j)}(t) + \bm{W}_{\bm{B}} u(t) u_j(t) \\
    y^{(j)} &= u^\top \bm{W}_{\bm{C}}^{\top} x^{(j)}(t).
    \label{eq:Mamba_siso}
\end{aligned}
\end{equation}
To derive a state-space model including all states and inputs, we stack the states in \eqref{eq:Mamba_siso} to get one single state vector $x = \begin{bmatrix} (x^{(1)})^\top \; \dots\; (x^{(d)})^\top \end{bmatrix}^{\top} \in \mathbb{R}^{Nd}$
that satisfies the following state-space model:
\begin{equation}
\begin{aligned}
    \dot{x}(t) &= \bm{A}_c x(t) + \bm{B}_c u(t) \\
    y(t) &= \bm{C}_c x(t)
    \label{eq:selective_full_continuous}
\end{aligned}
\end{equation}
in which
\begin{equation}
\begin{array}{c@{\quad}l}
\displaystyle
\bm{A}_c = \begin{bmatrix}
     \bm{A}^{(1)} & \bm{0} & \cdots & \bm{0} \\
    \bm{0} & \bm{A}^{(2)} & \cdots & \bm{0} \\
    \vdots & \vdots & \ddots & \vdots \\
    \bm{0} & \bm{0} &  \cdots & \bm{A}^{(d)}
\end{bmatrix},
&
\begin{aligned}
    \bm{B}_c(t) &= \mathbf{I}_d \otimes \bm{W_B} u(t) \\
    \bm{C}_c(t) &= \mathbf{I}_d \otimes u(t)^\top \bm{W_C}^\top
\end{aligned}
\end{array}
\label{eq:ABC_full}
\end{equation}
For the discretization step, we follow the official implementation of Mamba \cite{gu2024mamba}:
\begin{equation}
\label{eq:discretization_Mamba}
\begin{aligned}
    \bm{A}[t] &= \exp\bigl(\Delta(t) \bm{A}_c\bigr),
    \quad
    \vcenter{
        \hbox{
            $\begin{aligned}
                \bm{B}[t] &= \Delta(t) \bm{B}_c \\
                \bm{C}[t] &= \bm{C}_c
            \end{aligned}$
        }
    }
\end{aligned}
\end{equation}
which uses a zero-order hold (ZOH) for the matrix $\bm{A}$, and a simplified Euler discretization for the matrix $\bm{B}$. The matrix $\bm{C}$ remains the same as $\bm{C}_c$ since it represents a static relationship between the output $y$ and state $x$.
Utilizing this discretization procedure and the selective SSM architecture in \eqref{eq:selective_full_continuous} results in the following discrete-time state-space model:
\begin{equation}
\label{eq:selective_ssm}
    \begin{aligned}
        x[t+1] &= e^{\Delta[t] \bm{A}_c} x[t] + \Delta[t] \big( \mathbf{I}_d \otimes \bm{W_B} u[t] \big) u[t] \\
        y[t] &= \big( \mathbf{I}_d \otimes u[t]^\top \bm{W_C}^\top \big) x[t] \\
        \Delta[t] &= \tau_\Delta(p + q^\top u[t]).
    \end{aligned}
\end{equation}
Assuming $x(0) = 0$, we can unroll this recursive relation:
\begin{equation}
\label{eq:unrolled_io_selective}
    \begin{aligned}
        y[t'] &= \big( \mathbf{I}_d \otimes u[t']^\top \bm{W_C}^\top \big) \sum_{t=0}^{t'-1} \Big( \bm{A}^t \Delta[t'-1-t] \big( \mathbf{I}_d \otimes \bm{W_B} u[t'-1-t] \big) u[t'-1-t] \Big) \\
    \end{aligned}
\end{equation}
in which $\bm{A}^t$ is a shorthand notation for
\begin{equation}
\begin{aligned}
    \bm{A}^0 &= \mathbf{I}, \quad \bm{A}^t = e^{(\Delta[t'-1] + \cdots + \Delta[t'-t])\bm{A}_c} \text{ for } t>0.
    \label{eq:A^t_shorthand}
\end{aligned}
\end{equation}

For classification tasks requiring a scalar output, an additional parameter $w \in \mathbb{R}^d$ is introduced. This parameter corresponds to a linear layer applied to the last time step of the output sequence to obtain a label as $z=w^\top y[T]$, which captures all past information. The space of all selective SSMs $\mathcal{F}_{\text{SSM}}$ as defined in \eqref{eq:selective_ssm} is parametrized  by
\begin{equation}
    \Theta_{\text{SSM}} = \{\bm{A}_c, \bm{W_B}, \bm{W_C}, p, q, w \}.
    \label{eq:ssm_theta}
\end{equation}


\section{Generalization Bounds for Selective SSMs}
\label{sec:generalization}
To quantify generalization, we study the gap between the empirical training loss and the true expected loss over unseen data. Following standard results from statistical learning theory, we upper bound this generalization gap using the Rademacher complexity of the function class. Rademacher complexity of a function class $\mathcal{F}$ for a given dataset $S$, denoted as $\operatorname{Rad}(\mathcal{F},S)$, measures the average ability of a function class to fit random noise and serves as a data-dependent notion of capacity. For completeness, we include the formal definition of Rademacher complexity and the precise statement of the theorem that bounds the generalization gap in terms of it in Appendix~\ref{sec:input_dependent_gen_err_proof}.

Bounding the Rademacher complexity of complex function classes, such as foundation models, is challenging. A common approach is to break down the model into smaller components and analyze each of them separately by employing covering numbers. Covering numbers quantify the size of each component's function class by determining how many smaller subsets, or ``balls'', are needed to cover it. Formally, the covering number is defined as follows.
\begin{definition}[\textbf{Covering number}]
    Let $\mathcal{M}$ be a metric space with metric $\mu$. A subset $\mathcal{H} \subset \mathcal{M}$ is an $\epsilon$-cover for $\mathcal{M}$ if for every $h \in \mathcal{M}$, there exists $\hat{h} \in \mathcal{H}$ such that $\mu(h,\hat{h}) \leq \epsilon$. The covering number $\mathcal{N}(\mathcal{M}, \epsilon, \mu)$ is the lowest cardinality of an $\epsilon$-cover of $\mathcal{M}$.
    \label{def:covering_number}
\end{definition}

\begin{remark}[\textbf{Types of covering numbers}]
\label{rem:cover_types}
In our analysis, two distinct types of covering numbers are employed. The metric space $\mathcal{M}$ in Definition~\ref{def:covering_number} can be chosen as a collection of matrices equipped with the metric induced by a matrix norm. We deploy this definition to construct a cover for the parameters $\bm{A}_c$ and $p$, similar to the existing covering techniques developed for RNNs. On the other hand, let $\mathcal{F} = \{ f : \mathcal{U} \rightarrow \mathcal{Z} \}$ denote a function class, where $\mathcal{Z}$ is equipped with a norm $\|\cdot\|$, and let $S = \{u_{(i)}\}_{i=1}^m$ be a dataset. By equipping $\mathcal{F}$ with the following metric
\begin{equation}
\label{eq:data_metric}
    \mu_{p,\|\cdot\|}(f, \hat{f}) = \left( \frac{1}{m} \sum_{i=1}^m \left\| f(u_{(i)}) - \hat{f}(u_{(i)}) \right\|^p \right)^{1/p},
\end{equation}
we obtain a \emph{data-dependent} covering number for a function class, to construct covers for $\bm{W}_{\bm{B}}, \bm{W}_{\bm{C}}, q$, and $w$. This is parallel to the cover construction for the query, key, and value weight matrices in self-attention. For convenience, we denote the covering number of a function class $\mathcal{N}(\mathcal{F}, \epsilon, \mu_{p,\|\cdot\|})$ by $\mathcal{N}_p(\mathcal{F}_{|S}, \epsilon, \|\cdot\|)$.
\end{remark}

The covering numbers outlined in Remark~\eqref{rem:cover_types} can be utilized to bound the Rademacher complexity of a selective SSM parametrized as in \eqref{eq:ssm_theta} via Dudley’s integral theorem, stated below.
\begin{theorem}[\citet{bartlett2017spectrally}, Lemma A.5]
\label{thm:dudley}
    Given a real-valued function class $\mathcal{F} = \{ f:\mathcal{U} \rightarrow \mathbb{R} \}$ such that $\forall u \in \mathcal{U}, \; | f(u) | \leq \mathfrak{b}$ and a set of vectors $S = \{u_{(i)}\}_{i=1}^m$, we have
    \begin{equation}
        \begin{aligned}
            &\operatorname{Rad}(\mathcal{F}, S) \leq \inf_{\alpha > 0} \left( 4 \alpha + 12 \int_\alpha^{\mathfrak{b}} \sqrt{\frac{\ln \mathcal{N}_2 (\mathcal{F}_{|S}, \epsilon, \|\cdot\|_2)}{m}} \; d\epsilon \right).
        \end{aligned}
            \footnote{This is a modified version of Dudley's integral theorem \cite{dudley1967sizes}. The proof presented in \cite{bartlett2017spectrally} ignored the normalization by $m$ in \eqref{eq:data_metric} and takes $\mathfrak{b}=1$ in \eqref{eq:dudley}.}
        \label{eq:dudley}
    \end{equation}
\end{theorem}

\subsection{Main Result}
\label{sec:input_dependent_selective_scan}
In this section, we present an upper bound on the generalization error for selective SSMs. Before that, we outline and justify the assumptions necessary for deriving the bound. These assumptions are needed mostly to construct suitable covers through the lemmas presented in Appendix~\ref{sec:covering_numbers}.

\begin{assumption}[Input]
\label{ass:input_bound}
The input sequence $\|u[t]\|_2\le\mathfrak B_u$ for all $t$.
\end{assumption}

\begin{assumption}[Parameters]
\label{ass:parameter_bounds}
The parameters obey $\|\bm W_B\|_2\le\mathfrak B_{\bm B}$, $\|\bm W_B\|_{1,1}\le\mathfrak M_{\bm B}$, $\|\bm W_C\|_2\le\mathfrak B_{\bm C}$, $\|\bm W_C\|_{1,1}\le\mathfrak M_{\bm C}$, $\|w\|_2\le\mathfrak B_w$, $\|w\|_1\le\mathfrak M_w$, $\|q\|_2\le\mathfrak B_q$, and $\|q\|_1\le\mathfrak M_q$.
\end{assumption}

\begin{assumption}[Loss function]
\label{ass:loss_function}
The loss function $l(\cdot)$ is bounded by $\mathfrak c_l$ and Lipschitz continuous with constant $\mathfrak l_l$.
\end{assumption}

\begin{assumption}[State Matrix $\bm A_c$]
\label{ass:Ac_properties}
The continuous‐time state matrix $\bm A_c$ satisfies $\|\bm A_c\|_2\le\mathfrak B_{\bm A}$ and $\|\bm A_c\|_{2,1}\le\mathfrak M_{\bm A}$.
\end{assumption}

The bounds on the norms $\|\cdot\|_2$, $\|\cdot\|_1$, and $\|\cdot\|_{1,1}$ in Assumptions~\ref{ass:input_bound} and~\ref{ass:parameter_bounds} are standard and chosen to enable the use of Lemma~\ref{lem:linear_func_cover}, adapted from \cite{trauger2024length_independent_transformer} originally developed for Transformers. This result provides a covering number bound for linear function classes with bounded $\|\cdot\|_{1,1}$ norm, without introducing sample size or sequence length dependencies. While alternative norms can be used, they typically lead to looser or length-dependent bounds. The Lipschitz continuity of the loss function in Assumption~\ref{ass:loss_function} ensures that small changes in the input lead to controlled changes in the loss, which is essential for bounding the model's generalization error in Lemma~\ref{lem:generalization_error_bound_last}. Assumption~\ref{ass:Ac_properties} employs different norms than \ref{ass:parameter_bounds} ($\| \cdot \|_{2,1} $ rather than $\| \cdot \|_{1,1}$) because we cover $\bm A_c$ with a distinct strategy (Lemma~\ref{lem:cover_A_selective}) from the other parameters. With these in place, we now state the following theorem, whose proof is provided in Appendix~\ref{sec:input_dependent_gen_err_proof}.

\begin{theorem}
\label{thm:gen_err_bound_selective}
Let $S = \{u_{(i)}, z_{(i)}\}_{i=1}^m$ be the training set and $\mathcal{F}_{\text{SSM}}$ be all selective SSM blocks described in \eqref{eq:selective_ssm}. If Assumptions~\ref{ass:input_bound}--\ref{ass:Ac_properties} are satisfied, then with probability more than $1 - \delta$ the following bound for any $h \in \mathcal{F}_{\text{SSM}}$ holds:
\begin{equation}
    \begin{aligned}
        & \left| \mathbb{E}_{u,z}(l(h(u), z)) - \frac{1}{m} \sum_{i=1}^m l\left(h(u_{(i)}), z_{(i)}\right) \right| \leq \frac{12 \mathfrak{l}_l \mathcal{C}_{\mathcal{F}_{\text{SSM}}}}{\sqrt{m}} \left( 1 + \ln\left(\frac{\mathfrak{c}_l\sqrt{m}}{3 \mathcal{C}_{\mathcal{F}_{\text{SSM}}}}\right) \right) + 3\mathfrak{c}_l \sqrt{\frac{\ln\left(\frac{2}{\delta}\right)}{2m}}
    \end{aligned}
    \label{eq:gengap_fixed}
\end{equation}
in which
\begin{equation}
    \begin{aligned}
        \mathcal{C}_{\mathcal{F}_{\text{SSM}}} &= \mathcal{\tilde{O}}\left( \mathfrak{M}_\Delta \mathfrak{B}_w \mathfrak{B}_u^3 \mathfrak{B}_{\bm{B}} \mathfrak{B}_{\bm{C}} \mathfrak{B}_{\bm{A}} S_2 (\mathfrak{M}_\Delta^{2/3} N^{1/3} d^{1/3} + \mathfrak{B}_q^{2/3} \mathfrak{B}_u^{2/3} )^{3/2} \right)
    \end{aligned}
    \label{eq:capacity_input_dep}
\end{equation}
and
\begin{equation}
\label{eq:rho_Mdelta_S}
\begin{aligned}
    S_2 =
        \frac{\rho_{\bm{A}}(1 - \rho_{\bm{A}}^T)}{(1 - \rho_{\bm{A}})^2} - \frac{T\rho_{\bm{A}}^T}{\rho_{\bm{A}} - 1}, \quad \rho_{\bm{A}} = \left( 1 + e^{p - \mathfrak{B}_q \mathfrak{B}_u} \right)^{{s_{\bm{A}}}+\eta}, \quad \mathfrak{M}_{\Delta} &= \ln( 1 + e^{p + \mathfrak{B}_q \mathfrak{B}_u} ),
\end{aligned}
\end{equation}
\end{theorem}
where $\eta > 0$ is any arbitrarily small positive number chosen in a way that $s_{\bm{A}} + \eta \neq 0$. The notation $\mathcal{\tilde{O}}(\cdot)$ ignores the logarithmic dependencies on $N$ and $d$, but not $T$. The terms $ \mathfrak{M}_{(\cdot)} $ does not appear in the capacity expression $ \mathcal{C}_{\mathcal{F}_{\text{SSM}}} $, with the exception of $\mathfrak{M}_{\Delta}$. This is the result of an assumption $\mathfrak{M_{(\cdot)}}=\mathfrak{B}_{(\cdot)}$, made for the ease of presentation in the proof. To ensure clarity in the derivation, these terms are handled separately throughout the proof, and the assumption is incorporated only at the final stage.

\begin{proof}[Proof Sketch of Theorem \ref{thm:gen_err_bound_selective}]
The class of all Selective SSMs, denoted by $\mathcal{F}_{\text{SSM}}$, is parameterized by  
$\Theta_{\mathrm{SSM}}=\{\bm{A}_c, \bm{W_B}, \bm{W_C}, p, q, w\}$. We bound the generalization gap $|\mathbb{E} (l(h(u),z)) - \tfrac1m\sum_{i=1}^m l(h(u^{(i)}),z^{(i)})|$ by controlling the Rademacher complexity of $\mathcal{F}_{\text{SSM}}$.

The first challenge is to bound the distance between $\bm{A}^t$ and its cover $\hat{\bm{A}}^t$. Unlike LTI SSMs and RNNs, where the fixed number $\|\bm{A}\|_2^t$ bounds $\|\bm{A}^t\|_2$, here $\|\bm{A}^t\|_2$, as defined by the shorthand in \eqref{eq:A^t_shorthand}, has an input-dependent structure and must be controlled via the learnable parameters $\{p,q,\bm{A}_c\}$. In Lemma~\ref{lem:A^t_rho^t_bound_selective} we construct the appropriate $\rho_{\bm{A}}$ (as defined in \eqref{eq:rho_Mdelta_S}) using Gelfand’s formula (Corollary 5.6.14 in \cite{horn2012matrix}) and bounds on $p,q,\bm{A}_c$ to show $\|\bm{A}^t\|_2\le\rho_{\bm{A}}^t$, revealing the role of the spectral abscissa of $\bm{A}_c$. Gelfand’s formula characterizes the growth of $\| \bm{A}^t \|_2$ in terms of the spectral radius of $\bm{A}$, which is why the small positive number $\eta>0$ appears in \eqref{eq:rho_Mdelta_S}. Then Lemma~\ref{lem:A^t-Ahat^t_bound_selective} establishes an upper bound on $\|\bm{A}^t-\hat{\bm{A}}^t\|_2$ through a telescoping-sum argument. Finally, Lemma~\ref{lem:A_cover_selective}, followed by Lemma~\ref{lem:epsilon_sum_bound_selective}, shows how the cover radius for the whole model is decomposed as the sum of the cover radii of each parameter in $\Theta_{\mathrm{SSM}}$. Additionally, the structure of the input-dependent time step $\Delta[t]$ should be taken into account, as it affects both $\bm{A}[t]$ and $\bm{B}[t]$. The bounded norm assumptions on the input and $q$, together with the presence of the bias term $p$ in $\Delta[t]$, ensure the existence of a uniform lower bound on the step size. As shown in Lemma~\ref{lem:A^t_rho^t_bound_selective}, this bound is essential for the application of Gelfand's formula to control the growth of $\|\bm{A}^t\|_2$.

Next, individual covers are constructed: $\bm{A}_c$ is covered in Lemma~\ref{lem:cover_A_selective} as an element of the matrix space equipped with its matrix norm. Recognizing the attention mechanism in \eqref{eq:unrolled_io_selective}, Lemma~\ref{lem:linear_func_cover} covers the parameters $\{\bm{W}_B,\bm{W}_C,q,w\}$ by treating them as linear function classes. These individual covers are combined via a Cartesian product to cover $\Theta_{\mathrm{SSM}}$, and the cover radii are chosen optimally according to Lemma~\ref{lem:eq_constr_optim} to yield a global $\epsilon$‐cover of $\mathcal{F}_{\mathrm{SSM}}$. Finally, Lemma~\ref{lem:generalization_error_bound_last}, a standard corollary of the Dudley integral bound on Rademacher complexity, yields the claimed bound.
\end{proof}

\begin{remark}[\textbf{Lens of Attention}]
    A novel component of the proof of Theorem~\ref{thm:gen_err_bound_selective} lies in recognizing the attention mechanism in \eqref{eq:unrolled_io_selective}, which guides us to cover the parameter set $\{\bm W_B, \bm W_C, q, w\}$ as a function class. In particular, the line of work by \cite{zhang2002covering, edelman2022inductive, trauger2024length_independent_transformer} on generalization in Transformers presents numerous covering‐number bounds for such parameters. We utilize these results to establish our theorem, obtaining a length‐independent generalization bound under the condition $s_{\bm A}<0$. This result is attainable only by revealing the underlying attention mechanism. To show this connection explicitly, we make the following assumptions to single out the attention in selective SSMs:
\end{remark}

\begin{assumption}
\label{ass:constant_step_size}
Set $q = 0$ and $p = e$, yielding a constant step size $\Delta[t] = 1$.
\end{assumption}
\begin{assumption}
\label{ass:identity_state_matrix}
Assume $\bm{A}_c = \bm{0}$, so that $\bm{A}[t] = \mathbf{I}$ for all $t$.
\end{assumption}
Under Assumptions \ref{ass:constant_step_size} and \ref{ass:identity_state_matrix}, the selective SSM reduces to linear attention with a causal mask \cite{dao2024transformers_are_ssms}. In particular, the output at time $T$ admits the representation
\begin{equation}
\label{eq:unrolled_lin_attention_qkv}
z = w^\top 
\sum_{t=0}^{T-1}
\underbrace{\bigl(\mathbf{I}_d \otimes u[T]^\top \bm{W}_C^\top\bigr)}_{\text{Query}}
\;
\underbrace{\bigl(\mathbf{I}_d \otimes \bm{W}_B\,u[T-1-t]\bigr)}_{\text{Key}}
\;
\underbrace{u[T-1-t]}_{\text{Value}}.
\end{equation}
By fixing $\bm A[t]=\mathbf I$ and $\Delta[t]=1$, we obtain a significantly simpler derivation of the generalization error bound for linear attention given below, with its proof presented in Appendix~\ref{sec:lin_att_gen_err}.

\begin{proposition}
\label{prop:gen_error_linear_attention}
Let $S = \{u_{(i)}, z_{(i)}\}_{i=1}^m$, and suppose Assumptions~\ref{ass:input_bound}--\ref{ass:Ac_properties} hold with the simplifications made in Assumptions~\ref{ass:constant_step_size} and \ref{ass:identity_state_matrix}. Then, for the class of linear attentions $\mathcal{F}_{\text{LA}}$ parametrized by $\{\bm{W}_C, \bm{W}_B, w\}$, with probability at least $1-\delta$, the generalization bound in \eqref{eq:gengap_fixed} holds, where $\mathcal{C}_{\mathcal{F}_{\mathrm{SSM}}}$ is replaced by
\begin{equation}
    \mathcal{C}_{\mathcal{F}_{\text{LA}}}
    = \tilde{\mathcal{O}}\bigl( T \mathfrak{B}_{w} \mathfrak{B}_{\bm{B}} \mathfrak{B}_{\bm{C}} \mathfrak{B}_{u}^3 \bigr).
\end{equation}
\end{proposition}


\section{Analysis}
\label{sec:analysis}

In this section, we draw insights from Theorem~\ref{thm:gen_err_bound_selective}: we interpret the dependency of the bound on its parameters and compare it to similar bounds derived for other architectures. Table~\ref{tab:comp_gen_err_bounds} summarizes the dependency of the generalization error bound on the sequence length $T$, the input dimension $d$, and the input magnitude $\mathfrak{B}_u$ for different architectures.

\textbf{Length Independence.} As shown in Theorem~\ref{thm:gen_err_bound_selective} and Table~\ref{tab:comp_gen_err_bounds}, when the spectral abscissa satisfies $s_{\bm{A}}<0$, the generalization bound is length-independent, whereas if $s_{\bm{A}}>0$, the bound grows exponentially in $T$. This aligns with prior results for RNNs \cite{chen2019generalization_rnn, zhang2018stabilizing} and LTI SSMs \cite{racz2024length, wang2024stablessm}, which similarly emphasize the importance of the Lipschitz constant of the activation function and norm of the state matrix. In our setting, the Mamba construction \eqref{eq:discretization_Mamba} ensures that the stability properties of the continuous-time matrix $\bm{A}_c$ carry over to the discrete-time matrix $\bm{A}[t]$, directly controlling the generalization gap. 

\textbf{Attention.} We now compare the bounds we derived in Theorem~\eqref{thm:gen_err_bound_selective} for selective SSMs and Proposition~\eqref{prop:gen_error_linear_attention} for linear attention. The gap between their capacity terms, $\mathcal{C}_{\mathcal{F}_{\text{SSM}}}$ and $\mathcal{C}_{\mathcal{F}_{\text{LA}}}$, arises from simplifying assumptions used to reduce a selective SSM to a linear attention structure. First, Assumption~\ref{ass:constant_step_size} removes dependence on the input-conditioned discretization $\Delta[t]$, eliminating the $\mathfrak{M}_{\Delta}$ term and its associated input-norm $\mathfrak{B}_u$ scaling, resulting in a $\mathfrak{B}_u^3$ dependence—reflecting the three linear projections of attention (query, key, value). In contrast, selective SSMs incur an extra $\mathfrak{B}_u$ factor embedded in $\mathfrak{M}_{\Delta}$ due to dynamic input influence. Second, Assumption~\ref{ass:identity_state_matrix} further simplifies the bound by removing the need to cover the state matrix $\bm{A}$, reducing the $S_2$ term to a linear dependence on $T$ due to projection accumulation. Lastly, comparing linear to softmax attention, the key difference lies in normalization. Softmax reweighs the sequence, removing any explicit $T$-dependence in the bound and effectively decoupling capacity from sequence length.

\begin{table}[t]
\setlength{\tabcolsep}{9pt}
\centering
\fontsize{8}{8}\selectfont
\caption{\small{\textbf{Generalization bound scaling for different sequence-to-sequence models.} Dependencies are shown with respect to sequence length $T$, hidden dimension $d$, and input magnitude $\mathfrak{B}_u$ (logarithmic terms in $d$ and $\mathfrak{B}_u$ omitted). \dag \; The term $\rho_{\bm{A}}$ depends on the spectral abscissa $s_{\bm{A}}$, as defined in \eqref{eq:rho_Mdelta_S}. When $s_{\bm{A}} < 0$, we can choose $\eta > 0$ to be arbitrarily small so that $\rho_{\bm{A}} < 1$, yielding a length-independent bound. \ddag \; $\mathfrak{l}_x$ is the Lipschitz constant of the activation function used in the RNN as defined in \eqref{eq:RNN}. \S \; The bounds for LTI SSMs are derived under different assumptions as explained in Remark~\ref{rem:lti_ssms}.}}
    \begin{tabular}{l p{2.5cm} l l p{2.5cm}}
      \toprule
      \textbf{Model} & Specification & $T$ & $d$ & $\mathfrak{B}_u$ \\
      \midrule
      \multirow{2}{*}{Selective SSM (Theorem~\ref{thm:gen_err_bound_selective})}
        & $s_{\bm{A}} < 0$ & $1$ & $d^{1/2}$ & $\mathfrak{B}_u^4$ \\
        & $s_{\bm{A}} \geq 0$ & $T \rho_{\bm{A}}^T$ \textsuperscript{\dag} & $d^{1/2}$ & $\mathfrak{B}_u^4$ \\
        \midrule
        Linear Attention (Proposition~\ref{prop:gen_error_linear_attention} ) & N.A. & $T$ & $1$ & $\mathfrak{B}_u^3$ \\
      \midrule
      Softmax Attention \cite{trauger2024length_independent_transformer} & N.A. & $1$ & $1$ & $\mathfrak{B}_u^3$ \\
      \midrule
      \multirow{3}{*}{Vanilla RNN \cite{chen2019generalization_rnn}}
        & $\mathfrak{l}_x \|\bm{A}\|_2 < 1$ \textsuperscript{\ddag} & $1$ & $d$ & $\mathfrak{B}_u$ \\
        & $\mathfrak{l}_x \|\bm{A}\|_2 = 1$ & $T$ & $d$ & $\mathfrak{B}_u$ \\
        & $\mathfrak{l}_x \|\bm{A}\|_2 > 1$ & $T$ & $d^{3/2}$ & $\mathfrak{B}_u$ \\
      \midrule
      Discrete-time LTI SSMs \cite{racz2024length} \textsuperscript{\S} & $\max_i | \lambda_i(\bm{A}) | < 1 $ & $1$ & -- & Bounded $\sum_{k=1}^\infty \| u[k] \|_2^2 $ \\
      \midrule
      Continuous-time LTI SSMs \cite{liu2025autocorrelation} \textsuperscript{\S} & $s_{\bm{A}} < 0$ & $\ln^{3/2}(T)$ & -- & \small{Holder continuous} \\
      \bottomrule
    \end{tabular}
\vspace{-5pt}
\label{tab:comp_gen_err_bounds}
\end{table}

\textbf{RNNs.} Consider the vanilla RNN model  
\begin{equation}
\begin{aligned}
\label{eq:RNN}
x[t] = \sigma_x\bigl(\bm{A}\,x[t-1] + \bm{B}\,u[t]\bigr), \quad
y[t] = \sigma_y\bigl(\bm{C}\,x[t]\bigr),
\end{aligned}
\end{equation}
where $\sigma_x$ is $\mathfrak{l}_x$-Lipschitz and bounded. When $\mathfrak{l}_x\|\bm{A}\|_2 < 1$, the bound is length-independent, whereas for $\mathfrak{l}_x\|\bm{A}\|_2 \ge 1$ it exhibits linear dependence on $T$. The key mechanism RNNs avoid exponential dependence is the bounded activation \cite{cheng2024risk_rnn}. Unfortunately, incorporating a bounded activation into selective SSMs degrades their training efficiency, the very advantage that motivated their design. Another distinction is that RNN bounds hinge on contractivity---the induced norm of the matrix being less than one or related to the singular values---while our bound depends on the spectral properties of $\bm{A}_c$. This difference stems from our analysis, which begins with a continuous-time state-space model and discretizes via the matrix exponential, an operation governed by eigenvalues rather than singular values. We believe that the distinctive parameter for RNNs ($\mathfrak{l}_x \|\bm{A}\|_2$) can be improved in a similar way to our approach, where we took advantage of Gelfand’s formula. This allows replacing $\|\bm{A}\|_2$ with $\max_i |\lambda_i(\bm{A})|$, which is an improvement since the spectral radius of a matrix is a lower bound on any induced matrix norm.

\begin{remark}[\textbf{LTI SSMs}]
\label{rem:lti_ssms}
    One of the recent generalization bounds for LTI SSMs is the length-independent result of \citet{racz2024length}, which applies exclusively to stable discrete-time LTI systems. Their bound relies on the $\ell_1$-norm of the system's impulse response and the $\mathcal{H}_2$-norm of its transfer function, both of which are finite for only strictly stable systems. Extending this result to selective SSMs is nontrivial, as there is no direct analogue of these norms for general nonlinear systems. Moreover, they assume that the input satisfies $\sum_{k=1}^\infty \|u[k]\|_2^2 < \infty$, which is generally not satisfied in deep learning applications unlike our Assumption~\ref{ass:input_bound}. \citet{liu2024generalization} derive another generalization bound for LTI SSMs, but they aim to consider the temporal dependencies in the input signal via its mean and variance, leading to a Hölder continuity condition. Their bound, under a stability assumption, scales logarithmically with $T$ and is derived for continuous-time SSMs. Thus, in practice, it only applies when the discretized model remains close to its continuous-time counterpart. In contrast, we directly consider a continuous-time SSM with an input-dependent time scale, and derive a generalization bound for the resulting discrete-time model—an assumption more aligned with how such models are used in practice. In summary, while existing techniques for SSMs leverage system-theoretic properties unique to LTI systems, the nonlinearity of selective SSMs requires  deriving bounds that are better suited to modern deep learning architectures.
\end{remark}

\begin{figure*}[t]
    \centering

    \begin{minipage}[t]{0.31\textwidth}
        \centering        \includegraphics[width=\linewidth]{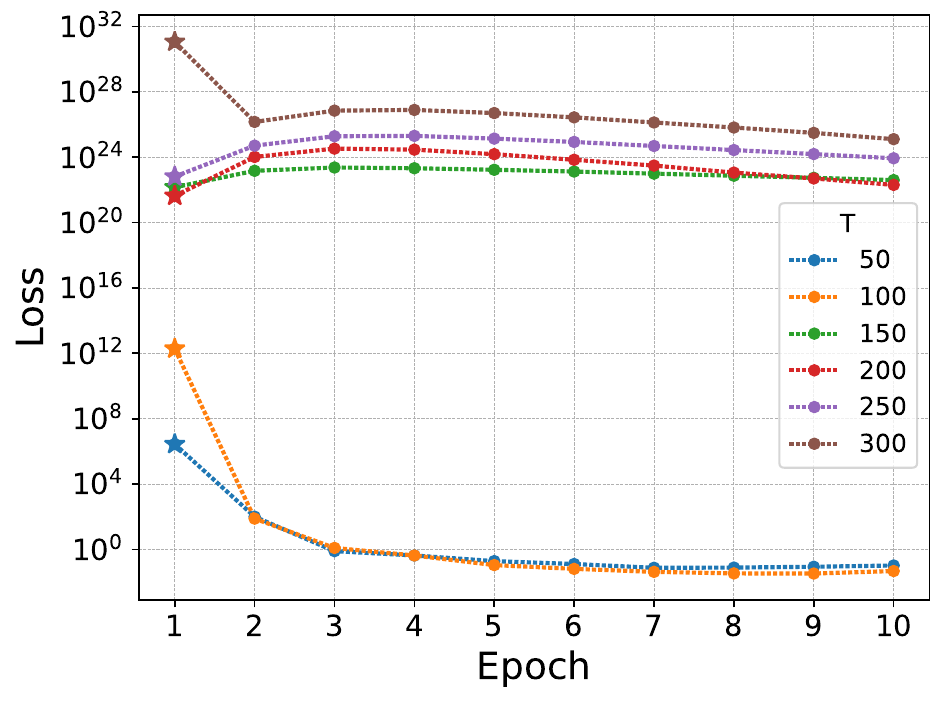} 
    \end{minipage}
    \begin{minipage}[t]{0.31\textwidth}
        \centering
        \includegraphics[width=\linewidth]{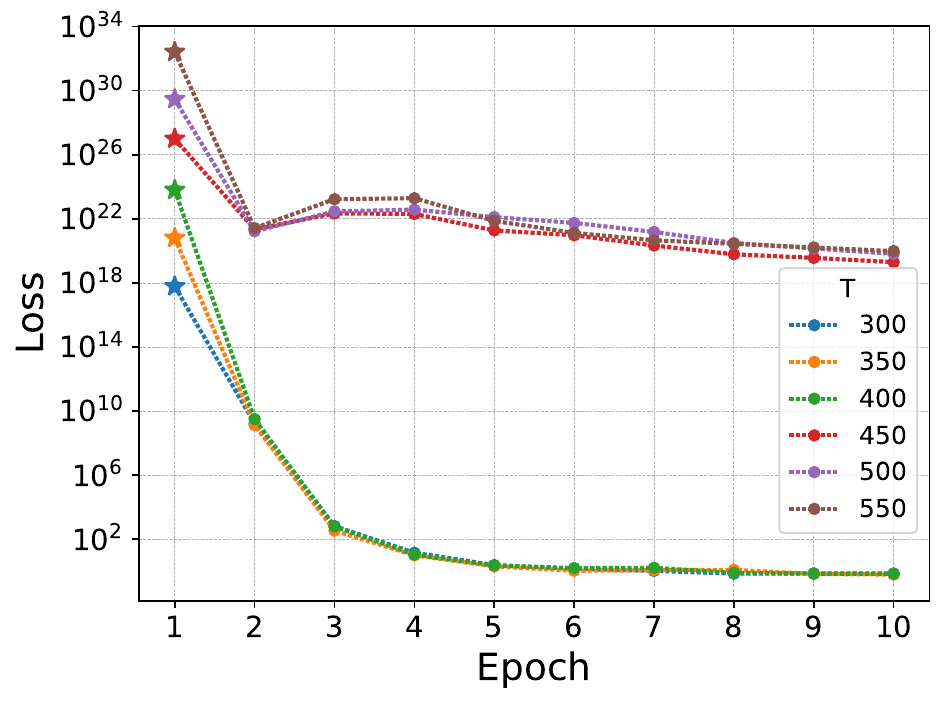} 
    \end{minipage}
    \begin{minipage}[t]{0.31\textwidth}
        \centering
        \includegraphics[width=\linewidth]{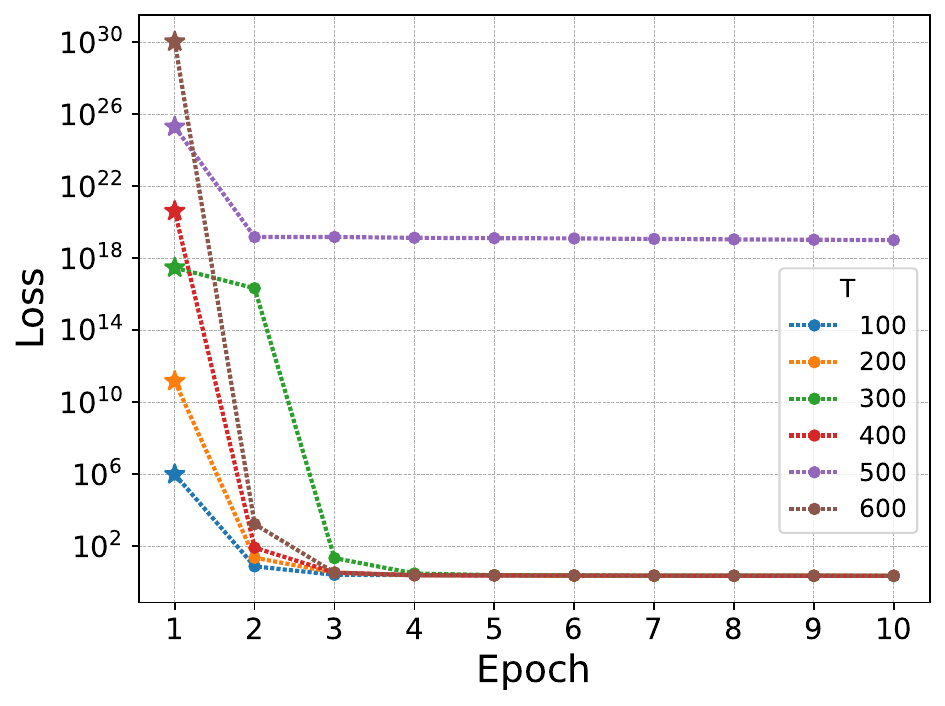} 
    \end{minipage}

    \vspace{0.8em} 

    \begin{minipage}[t]{0.32\textwidth}
        \centering
        \includegraphics[width=\linewidth]{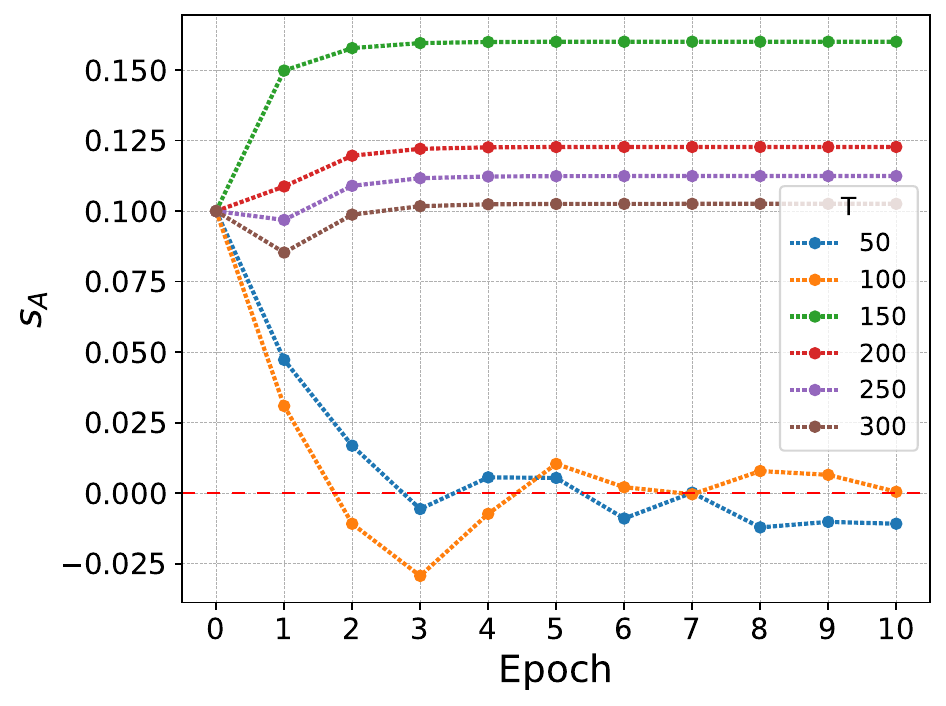} 
    \end{minipage}
    \begin{minipage}[t]{0.32\textwidth}
        \centering
        \includegraphics[width=\linewidth]{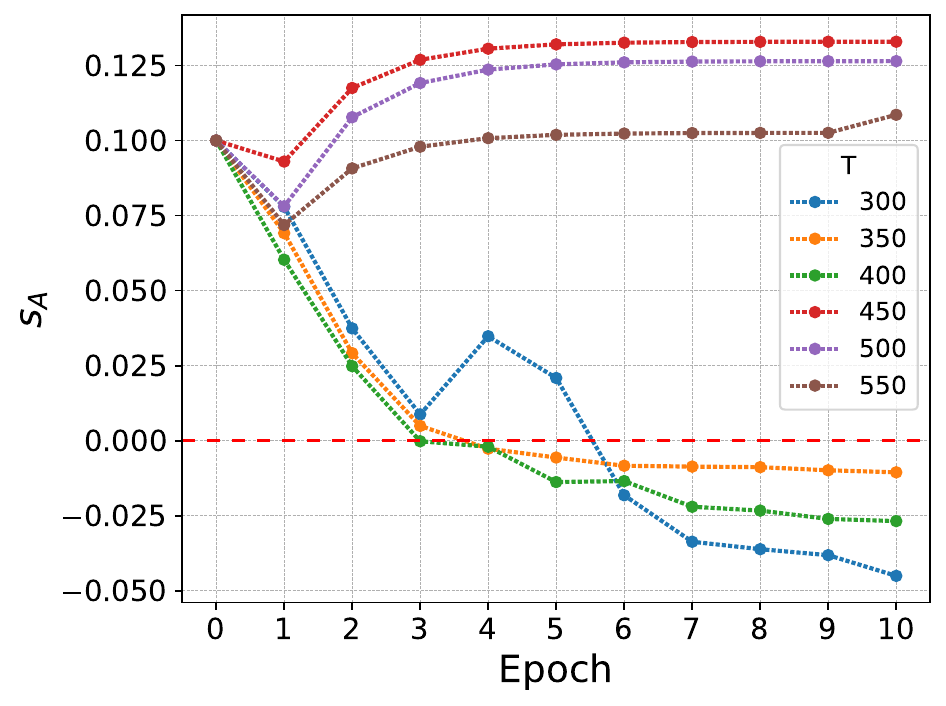} 
    \end{minipage}
    \begin{minipage}[t]{0.32\textwidth}
        \centering
        \includegraphics[width=\linewidth]{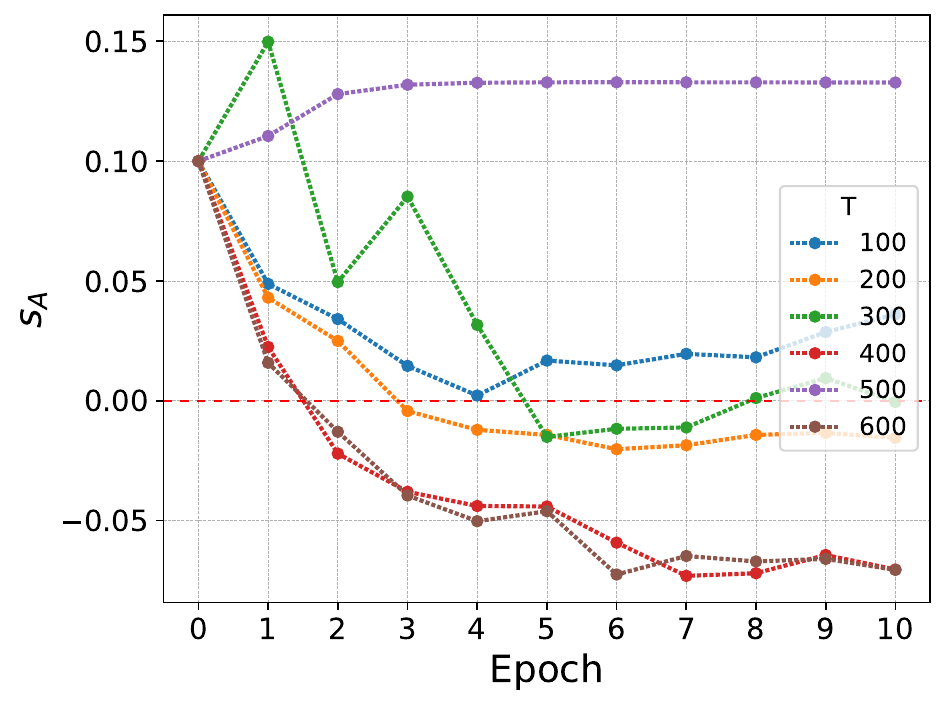} 
    \end{minipage}

    \caption{\small{\textbf{Experiment 1.}
    \textbf{Top:} Training loss vs epochs for \textbf{Left:} Majority, \textbf{Middle:} IMDb, \textbf{Right:} ListOps.
    \textbf{Bottom:} Evolution of $s_{\bm{A}}$ vs epochs for the same datasets.
    All runs use an \emph{unstable initialization} with $s_{\bm{A}}=0.1$.
    Whenever training successfully reduces the loss, the spectral abscissa $s_{\bm{A}}$ is driven toward zero, indicating that the system becomes stable. In cases where $s_{\bm{A}}$ does not decrease toward zero, training is not successful.}}
    \label{fig:unstable_behavior}
\end{figure*}

Overall, our results complete the picture of generalization for deep sequence models. Strongest generalization occurs in selective SSMs with $s_{\bm{A}} < 0$, softmax-attention, and RNNs with $\mathfrak{l}_x\|\bm{A}\|_2 < 1$, all benefiting from implicit normalization or stabilization. In contrast, selective SSMs with $s_{\bm{A}} > 0$ generalize poorly. We revisit this in the next section, showing that no sub-exponential bound can be derived using Rademacher complexity in this regime. Later on in Section~\ref{sec:experiments}, we observe that the training error grows rapidly unless $s_{\bm{A}}$ is driven negative, as seen in Figure~\ref{fig:unstable_behavior}.

\subsection{A Lower Bound on the Rademacher Complexity}
In this section, we present a theorem that establishes a lower bound on the Rademacher complexity for the case where the spectral abscissa satisfies $s_{\bm{A}} \geq 0$.
\begin{theorem}
\label{thm:lower_bound}
Let $s_{\bm{A}} > 0$ be a fixed spectral abscissa and $\mathcal{S} = \{u_{(i)}\}_{i=1}^{m} \subset [-1,1]^T$. If $|w| \leq \mathfrak{B}_w$, then
\begin{equation}
\operatorname{Rad}_{\mathcal{S}}(\mathcal{F}_{\text{SSM}}) \ge \mathfrak{B}_w \frac{(1 + s_{\bm{A}})^T - 1}{s_{\bm{A}}} \sqrt{\frac{2}{\pi m}}.
\end{equation}
Moreover, if $s_{\bm{A}} = 0$, then
\begin{equation}
\operatorname{Rad}_{\mathcal{S}}(\mathcal{F}_{\text{SSM}}) \ge \mathfrak{B}_w T \sqrt{\frac{2}{\pi m}}.
\end{equation}
\end{theorem}

The proof of the theorem is provided in Appendix~\ref{app:lower_bound}. It is based on the construction of a restricted subclass of selective SSMs in which the step size is fixed and the output grows unboundedly as a function of a single parameter \(w\). In this simplified setting, the Rademacher complexity can be computed explicitly, yielding a lower bound on the Rademacher complexity of the broader class \(\mathcal{F}_{\mathrm{SSM}}\). Unfortunately, these lower bounds are looser than the corresponding upper bounds by a factor of \(\mathcal{O}(T)\), suggesting that there is room for improvement in at least one of the bounds. Nevertheless, they demonstrate an important fact: \textbf{the dependence on $T$ in the Rademacher complexity cannot be eliminated} when $s_{\bm{A}}\geq0$. Specifically, it must grow at least exponentially when $s_{\bm{A}}>0$.

\begin{figure*}[t]
    \centering
    \includegraphics[width=0.32\textwidth]{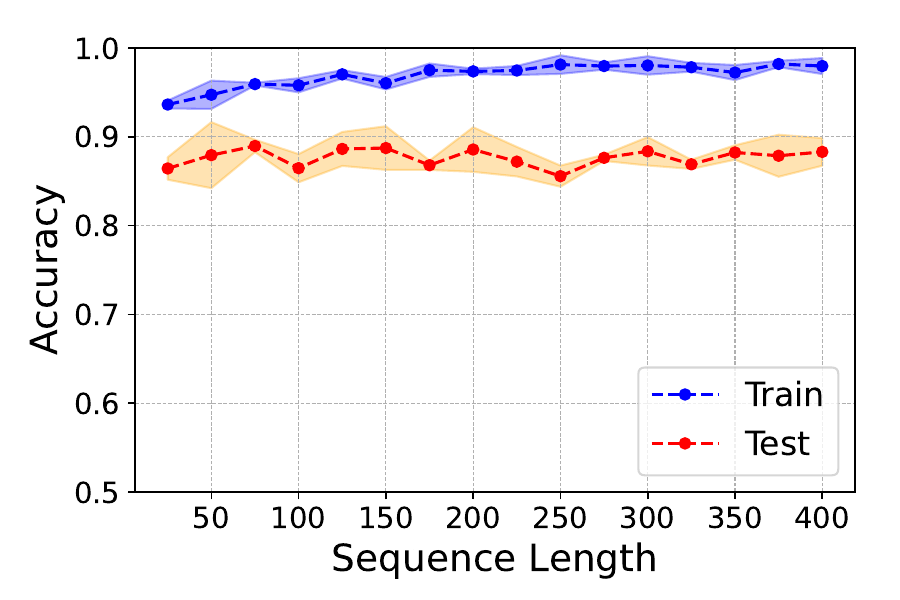}
    \hfill
    \includegraphics[width=0.32\textwidth]{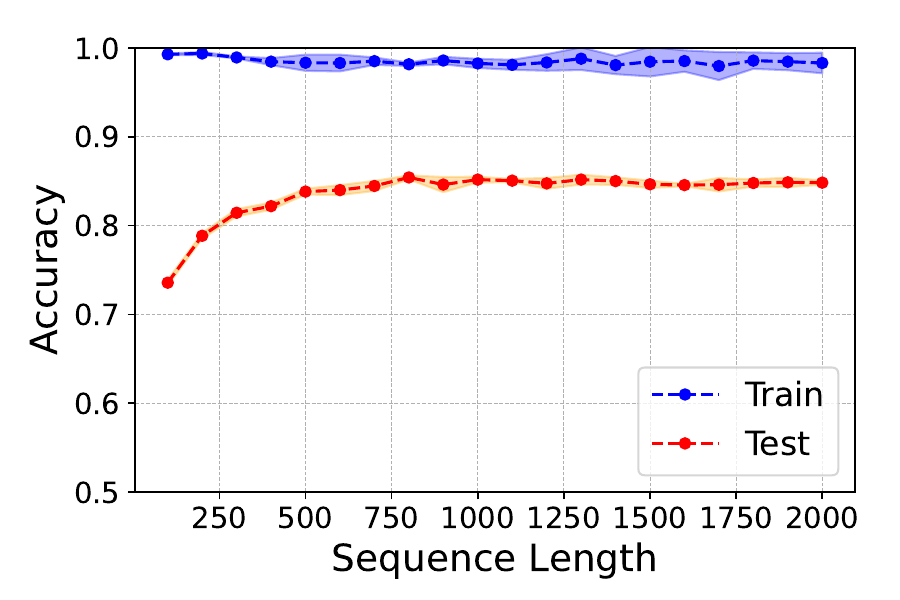}
    \hfill
    \includegraphics[width=0.32\textwidth]{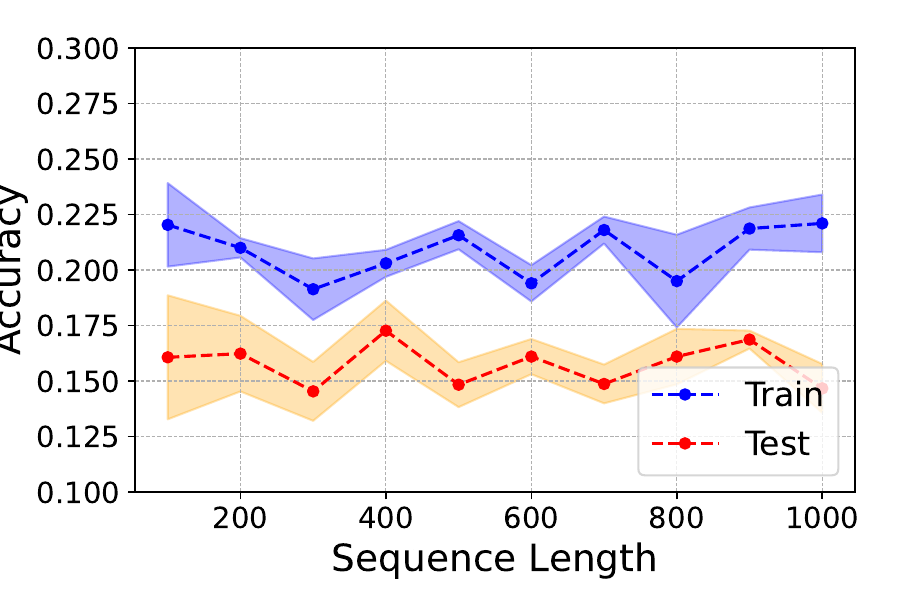}
    \caption{\small{\textbf{Experiment 2.}
    \textbf{Left:} Majority, \textbf{Middle:} IMDb, \textbf{Right:} ListOps. Train and test accuracy versus sequence length $T$ for models initialized with $s_{\bm{A}}=0$. The results demonstrate length-independent generalization. Each experiment is repeated five times with different random seeds; the dashed line denotes the mean accuracy across runs, and the shaded region represents $\pm$ one standard deviation.}}
    \label{fig:train_test_error}
\end{figure*}

\section{Experiments}
\label{sec:experiments}

To validate our theoretical findings, we conduct two sets of experiments on three datasets \footnote{Our code is available at \url{https://github.com/Arya-Honarpisheh/gen_err_sel_ssm}.}. The first experiment examines the effect of a positive spectral abscissa. It demonstrates that when the system matrix is initialized with positive $s_{\bm{A}}$, training may fail, especially for longer sequences, rendering the generalization gap ill-defined. Since this gap is defined as the difference between train and test error, if one of these two terms is not well-defined, it would be meaningless to evaluate their difference. In the second experiment, we initialize the state matrix with $s_{\bm{A}} = 0$ and study how the generalization gap evolves with increasing sequence length. The model architecture used in the experiments consists of an embedding layer, followed by a single selective SSM block. The model is trained using cross-entropy loss. To stabilize training, we employ a regularization function from \citet{keller2024regularization}. Below, we describe the tasks with their corresponding datasets, followed by our analysis of each experiment. 

\textbf{Majority:} The first task employs a synthetic “majority” dataset, where each input sequence consists of binary symbols $\{0,1\}$, embedded into a $d$-dimensional vector. The objective is to predict whether the sequence contains more ones than zeros. The output depends uniformly on all input positions: no single position has disproportionate influence, and the prediction is not determined by a small subset of the input sequence. This makes the task well-suited for measuring trends in generalization gap w.r.t. the sequence length $T$. During training, noise is introduced by randomly flipping a small percentage of the inputs after labeling, adding a layer of difficulty. The noise limits the model's accuracy around $95\%$, preventing it from overfitting despite the simplicity of the task.

\textbf{IMDb:} The second task is binary sentiment classification using the IMDb large movie review dataset \cite{maas2011imdb} containing 50K reviews. Each review is labeled as positive or negative based on its sentiment. This task poses a real-world challenge due to its high variability in sequence lengths and the need for contextual understanding. To control sequence length during training and evaluation, we pad/truncate sequences to a fixed $T$. For shorter sequences ($T\leq300$), sentiment indicators are often clear early, aiding prediction, while longer sequences often require retaining more context for accuracy. Thus, truncating them leads to a substantial decrease in model performance, as seen in the test loss in Figure~\ref{fig:train_test_error}. The average review length in the dataset is around $T=300$. Therefore, the test loss and generalization gap both stabilize after $T=300$, indicating that enough information is preserved for the model to generalize effectively. This is explained thoroughly in Appendix~\ref{sec:experimental_details_imdb}.

\textbf{ListOps:} The third task uses the ListOps dataset \cite{nangia2018listops}, a synthetic benchmark that evaluates a model’s ability to reason over hierarchical sequences. Each input is a bracketed expression with nested operations, for example \texttt{[MIN 5 1 [MAX 2 9] 0]}, which evaluates to a single-digit integer. The challenge lies in the fact that the correct output depends on the entire nested structure rather than local context. We use the version from \citet{tay2021long_range_arena} to align with standard long-sequence benchmarks.

\textbf{Experiment 1 (Stability Under Training):}  
This experiment investigates the behavior of the spectral abscissa $s_{\bm{A}}$ during training. To better understand the role of stability in training selective SSMs, we deliberately initialize all models in an \emph{unstable} regime with $s_{\bm{A}} = 0.1$. The key observation here is that successful training is accompanied by $s_{\bm{A}}$ being driven toward zero. We observe that as the sequence length $T$ increases, the initial loss grows exponentially, making it increasingly difficult for the model to escape instability. This ultimately causes training to fail for longer sequences. Since successful training consistently corresponds to $s_{\bm{A}}$ approaching zero, we consider the regime with $s_{\bm{A}} < 0$ as the stable operating region when analyzing the generalization gap in the next experiment. These results are illustrated in Figure~\ref{fig:unstable_behavior}.

\textbf{Experiment 2 (Length-Independent Generalization):}  
Here we evaluate the generalization behavior across varying sequence lengths. Unlike Experiment 1, we initialize the models in a \emph{marginally stable} regime with $s_{\bm{A}} = 0$, which results in smoother training. The models are trained and tested on sequences of different lengths, and we measure the generalization gap as the difference between training and test losses. As shown in Figure~\ref{fig:train_test_error}, this gap remains relatively stable across sequence lengths, with no consistent increasing or decreasing trend. These results support our theoretical claim that selective SSMs exhibit length-independent generalization behavior. 

Interestingly, the mechanism behind this behavior is foreshadowed in Experiment 1: when initialized in an unstable regime, the model naturally pushes $s_{\bm{A}}$ below zero during training, moving toward stability to enable learning. However, it stabilizes just enough to preserve rich temporal information. This suggests that \textbf{selective SSMs are implicitly biased toward operating near the stability boundary}, where they can extract long-range dependencies without incurring the exponential instability associated with longer sequences. This is a manifestation of the trade-off between expressivity and generalization. In Experiment 2, starting near this boundary allows the model to train smoothly, leading to consistent generalization across different sequence lengths.

\section{Conclusion and Future Work}

In this paper, we derived new generalization gap bounds for selective SSMs by leveraging their embedded linear‐attention mechanism. As a corollary to our main result, we obtained a bound for linear attention, which illustrates the underlying connections explicitly. Our analysis revealed that the spectral abscissa $s_{\bm{A}}$ of the continuous‐time state matrix $\bm{A}_c$ governs selective SSMs’ generalization behavior: models with $s_{\bm{A}} < 0$ enjoy length-independent guarantees, while those with $s_{\bm{A}} > 0$ suffer exponential growth in error. Finally, our experiments supported these theoretical findings, showing that models satisfying $s_{\bm{A}}<0$ indeed generalize well on long inputs. An important direction for future work is to improve these generalization error bounds. This could involve refining the analysis under alternative assumptions or exploring different techniques, such as directly bounding the Rademacher complexity instead of relying on covering-based arguments. Another promising avenue is to extend generalization analysis to other variants of deep architectures.

\section{Limitations}
This work considers a selective SSM as the discretization of a continuous-time state-space model, following the formulation used in Mamba. Hence, the theory developed here does not directly apply to other variants of selective SSMs where, for example, the matrix $A(\bm{u})$ exhibits a different dependency on the input. This work assumes that the training and test data are drawn from the same distribution. Out-of-distribution generalization of selective SSMs is not addressed. For a recent analysis of length generalization, a specific type of out-of-distribution setting where test sequences are longer than those seen during training, the interested reader is referred to \citet{buitrago2025len_gen}.

\section{Acknowledgments}

This work was partially supported by NSF grants CNS–2038493
and CMMI–2208182, AFOSR grant FA9550-19-1-0005, and ONR grant
N00014-21-1-2431.



\bibliographystyle{unsrtnat}
\bibliography{references}



\newpage

\appendix
\section{Related Work}

\textbf{Self-attention} is the core mechanism behind the Transformer architecture, introduced as an alternative to RNNs and CNNs for sequence processing \cite{vaswani2017attention, bahdanau2015neural}. While the concept of attention predates the Transformer, its introduction marked a pivotal shift in the scale of large models. An attention mechanism assigns scores to each pair of elements in a sequence to measure their relevance to each other. The self-attention mechanism, drawing inspiration from the key–query analogy used in relational databases, captures dependencies between elements of an input sequence, where inputs attend to each other within the same sequence. Since their introduction, Transformers have been extensively studied and refined, leading to numerous variants, including sparse and low-rank adaptations and widespread applications across domains such as natural language processing \cite{devlin2018bert, brown2020gpt3} and computer vision \cite{dosovitskiy2020vision_transformer, peebles2023scalable_diffusion_transformer, liu2024diffusion_vision_transformer}. Related to our work is the connection between SSMs and attention \cite{dao2024transformers_are_ssms, ali2024hidden}.

\textbf{State-space models} are a new class of foundation models, introduced by \citet{gu2021ssm} as an alternative to Transformers for sequence processing. Rooted in the classical state-space representations introduced by \citet{kalman1960new} in control theory, SSMs leverage state-space representations to efficiently model long-range dependencies in sequential data. The foundation of SSMs can be traced to the HiPPO framework, which established a mathematical basis for encoding and preserving long-range dependencies using orthogonal polynomial projections \cite{gu2020hippo}. Building on this foundation, the first practical implementation of SSMs is the S4 model, which utilized HiPPO as an initialization scheme \cite{gu2022s4}. With the empirical success of S4 on the Long Range Arena benchmark \cite{tay2021long_range_arena}, SSMs gained widespread attention, prompting several extensions and refinements. S4D simplified training with diagonal initializations \cite{gu2022s4d}, S5 introduced a multi-input multi-output structure for greater flexibility \cite{smith2023s5}, and Hyena explored hierarchical convolutions \cite{poli2023hyena}. Selective SSMs introduced in the Mamba model by \citet{gu2024mamba} extend LTI SSMs by using linear projections of the input to construct and discretize SSMs, resulting in a nonlinear time-variant architecture. These properties make selective SSMs closely resemble self-attention, as highlighted by \citet{dao2024transformers_are_ssms} while introducing Mamba-2. 

\textbf{Generalization bounds} are central in the probably approximately correct (PAC) learning framework, which formalizes a model's ability to achieve low error on unseen data with high probability, provided sufficient training data. The PAC–Bayes framework provides probabilistic guarantees on generalization through a KL divergence between posterior and prior distributions \cite{alquier2024pacbayes, mcallester1998some}. In the information-theoretic framework, the generalization gap is controlled by the mutual information between training data and model parameters \cite{xu2017information, steinke2020reasoning, haghifam2020sharpened}. Earlier studies explored statistical guarantees based on VC-dimension and shattering bounds extensively \cite{karpinski1997polynomial, koiran1997neural, sontag1998learning, baum1988size, bartlett1998almost}. A related line of work uses Rademacher complexity to control the generalization gap, often through chaining and Dudley’s integral, leading to margin- or norm-based bounds \cite{dudley1967sizes, vershynin2018high, shalev2014understanding, bartlett1998size, bartlett2017spectrally}. This framework was later refined through local Rademacher complexity \cite{bartlett2005local}, which focuses on subsets of the hypothesis class near the empirical minimizer, yielding sharper bounds for deep models \cite{wei2019data}. The recent works on norm-based generalization bounds utilize covering numbers, a fundamental tool for bounding the capacity of function classes \cite{vershynin2018high}. \citet{zhang2002covering} laid the groundwork for understanding the capacity of regularized linear function classes through covering numbers. Later, \citet{bartlett2017spectrally} established generalization bounds for neural networks using covering numbers based on the work of \citet{zhang2002covering}. These methods have been extended to Transformers in recent studies \cite{edelman2022inductive, trauger2024length_independent_transformer, truong2024rank}, where different aspects such as length or rank dependency have been emphasized. For LTI SSMs, the works of \citet{racz2024length, liu2024generalization} draw inspiration from this line of research but primarily leverage the structure of LTI systems to derive their bounds from a state-space perspective.

\section{Experimental Details}
\label{sec:experimental_details}

In both experiments, we employ an embedding layer implemented using \texttt{torch.nn.Embedding}, which maps input tokens into a continuous vector space. This is followed by a selective SSM block, configured with $N = 4$ states per channel and $d = 16$ channels. The selective SSM block is parameterized as  
\begin{equation*}
\Theta_{\text{SSM}} = \{ \bm{A}_c, \bm{W_B}, \bm{W_C}, p, q, w \}
\end{equation*}  
where each component is defined in Section~\ref{sec:ssm}. The matrix $\bm{A}_c\in\mathbb{R}^{Nd\times Nd}$ represents the state matrices across channels. In the code implementation, we store $\bm{A}_c$ structured as a $\mathbb{R}^{d \times N}$ matrix where each row of $\bm{A}_c$ corresponds to the diagonal elements of a distinct diagonal state matrix $\bm{A}_{c}^{(j)} \in \mathbb{R}^{N \times N}$ for channel $j$, where $j \in \{1, \dots, d\}$. This parameterization follows the official implementation of Mamba~\cite{gu2024mamba}, ensuring computational efficiency while maintaining expressive capacity. The remaining parameters in $\Theta_{\text{SSM}}$ have the exact dimensions described in Section~\ref{sec:ssm}: $\bm{W_B}, \bm{W_C} \in \mathbb{R}^{N \times d}$, $q \in \mathbb{R}^{d}$, and $p \in \mathbb{R}$. The first experiment investigates how the stability margin affects training, particularly showing that the initial loss grows exponentially with sequence length $T$ when $s_{\bm{A}}>0$. To focus on early training dynamics, we train for only 10 epochs using 10\% of the dataset, balanced across labels, to control loss magnitude and avoid label bias. In contrast, the second experiment uses the full datasets to train models to convergence and evaluate the generalization gap. The second experiment is repeated five times with different random seeds. For very long sequences in the IMDb dataset, even with \( s_{\bm{A}} = 0 \) initialization, training may fail even at the very first stage, preventing the model from stabilizing during the first few epochs. In such cases, we retry with a different seed that yields a non-NaN value in the first epoch. The final results are reported as the mean over five successful runs $\pm$ one standard deviation.

\subsection{Majority Dataset}
\label{sec:experimental_details_majority}

\begin{figure}[ht]
    \centering
    \includegraphics[width=0.5\textwidth]{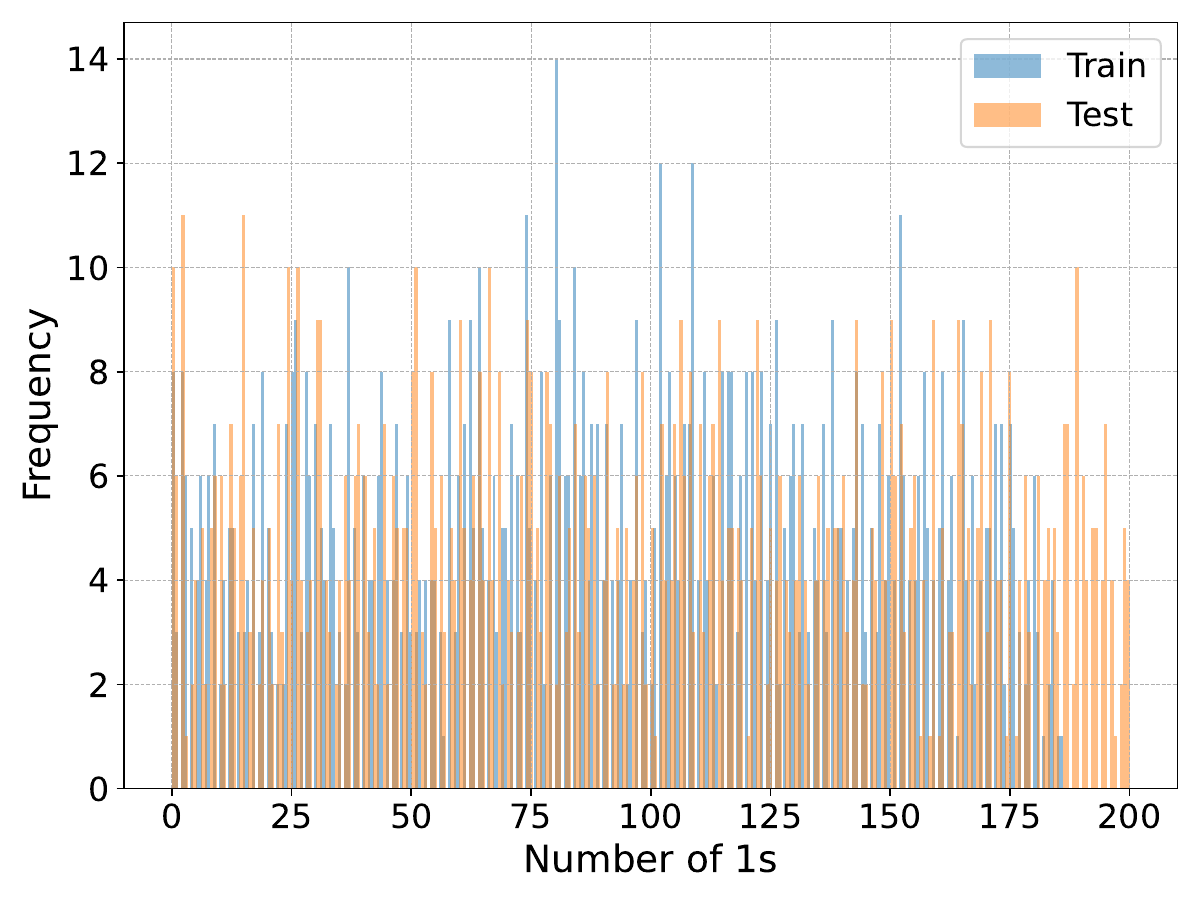}
    \caption{\textbf{Majority}. Histogram of ones, $m=1000$ samples each for train and test, sequence length $T=200$.}
    \label{fig:majority_ones}
\end{figure}

\vspace{8pt}

Majority is a synthetic dataset similar to the dataset used in the experiments of \citet{trauger2024length_independent_transformer}, but with modifications. Each sample consists of a sequence of ones and zeros, forming the basis of a binary classification task. The class label indicates whether a sequence contains more ones than zeros. A sample sequence $u_1$ with $T=20$ and its label $z_1$ would be as the following:
\begin{equation*}
    \begin{aligned}
        u_1 &= [1, 1, 0, 1, 1, 0, 1, 1, 0, 1, 1, 0, 1, 1, 0, 1, 0, 1, 0, 1] \\
        z_1 &= 1
    \end{aligned}
\end{equation*}

Since the task involves only two unique elements, the vocabulary size is set to 2, and each element in the sequence is projected into embeddings of dimension $d$ when they pass the embedding layer. Both the training and test sets contain $m = 1000$ samples. To ensure a uniform distribution of ones and zeros across sequence lengths, we generate sequences such that the number of ones varies approximately evenly from 0 to $T$. To introduce some imbalance, we modify the training set by randomly flipping $10\%$ of ones to zeros after generating the sequences and labels. As shown in Figure~\ref{fig:majority_ones}, this results in a noticeable reduction in sequences with a high number of ones. Specifically, towards the maximum sequence length $T$, fewer samples retain exactly $T$ ones due to these perturbations, altering the original distribution.


\subsection{IMDb Large Movie Review Dataset}
\label{sec:experimental_details_imdb}

\begin{figure}[ht]
    \centering
    \includegraphics[width=0.5\textwidth]{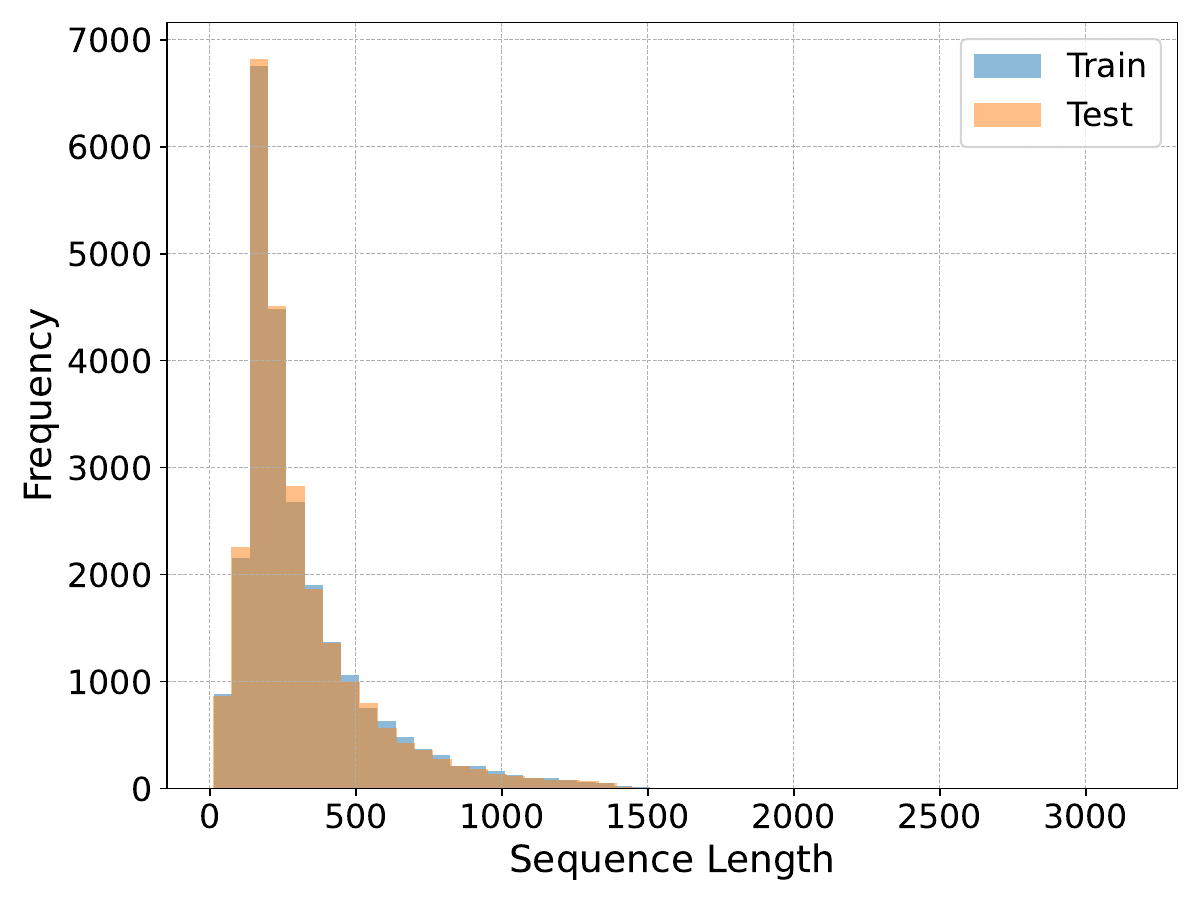}
    \caption{\textbf{IMDb}. Histogram of sequence lengths for both the training and test splits.}
    \label{fig:imdb_seq_lengths}
\end{figure}

\vspace{8pt}

IMDb large movie review dataset \cite{maas2011imdb} is a standard benchmark for sentiment analysis models and part of the Long-Range Arena (LRA) benchmark \cite{tay2021long_range_arena}. The dataset contains 50,000 movie reviews, evenly split between positive and negative labels, and is divided into training and test sets of 25,000 reviews each. The task is binary sentiment classification, aiming to predict whether a review expresses a positive or negative sentiment. The dataset's balanced nature ensures unbiased model evaluation.

We chose IMDb for its diverse sequence lengths, as shown in Table~\ref{tab:imdb_table} and Figure~\ref{fig:imdb_seq_lengths}. To train effectively, we used the entire dataset, truncating or padding sequences to a fixed length. For our experiments, we chose sequence lengths between 100 and 2000 tokens, based on the distribution observed in Figure~\ref{fig:imdb_seq_lengths}. As shown in Figure~\ref{fig:train_test_error} (bottom), test accuracy increases from 100 to 300 tokens, then stabilizes. The generalization gap, visible in the bottom plot, reflects this trend. The average sequence length is 314 tokens, with many sequences exceeding 300 tokens (Figure~\ref{fig:imdb_seq_lengths}). Truncating sequences longer than 300 tokens results in the loss of valuable information, potentially reducing predictive accuracy, as demonstrated by the following examples.
\begin{table}[h!]
\centering
\begin{tabular}{ll}\\
  \multicolumn{2}{l}{\textbf{Short Sample}}\\
  \textbf{Text:}   & ``I don't know why I \textcolor[rgb]{0.0, 0.5, 0.0}{like this movie} so well, but I never get tired of watching it." \\
  \textbf{Label:}  & Positive (1) \\
  \textbf{Length:} & 24 \\
  [1em]
  \multicolumn{2}{l}{\textbf{Long Sample}}\\
  \textbf{Text:}   & ``This movie was recently released on DVD in the US and I finally got the chance..." \\
  \textbf{Label:}  & Negative (0) \\
  \textbf{Length:} & 1833
\end{tabular}
\end{table}

For shorter sequences, key indicators of the sentiment label often appear early in the text, making it easier for the model to make predictions. However, for longer sequences, these indicators may not be immediately apparent, as the sentiment may be spread across the entire review. In such cases, retaining the full context of the sequence becomes crucial for accurate prediction. This is particularly evident in the test loss observed in the bottom Figure~\ref{fig:train_test_error}, where truncating longer sequences results in a loss of critical context, reducing the model's accuracy.

\vspace{8pt}

\begin{table}[h!]
\centering
\begin{tabular}{lcccc}
\toprule
       & Max   & Min & Average & Median \\ 
\midrule
Train  & 3127 & 13 & 314 & 233    \\ 
Test   & 3157 & 10 & 307 & 230    \\ 
\bottomrule
\end{tabular}
\vspace{8pt}
\caption{\textbf{IMDb}. Sequence length details for training and test splits.}
\label{tab:imdb_table}
\end{table}

\subsection{ListOps Dataset}
ListOps is a synthetic benchmark designed to evaluate a model's ability to perform hierarchical reasoning over long sequences first introduced in \citet{nangia2018listops}. Each sample in this dataset is a bracketed expression consisting of nested mathematical operations, such as
\[
[\text{MED}\; 4\; 8 \; [\text{MIN}\; 7\; 2]\; 2 \; 3],
\]
which evaluates to a single-digit integer. The correct label depends on the complete nested structure, making the task dependent on the entire length of the sequence. We adopt the preprocessed version provided by \citet{tay2021long_range_arena}. The vocabulary consists of digits, "[","]","(",")", and operator tokens (\texttt{MAX}, \texttt{MIN}, \texttt{MED}, \texttt{SUM}). Sequence lengths are set between $T = 100$ and $T = 1000$ in increments of 100. For each sequence length, we generate data in the range $[100k - 5,\, 100k + 5]$.

\subsection{Experiment 1: Bounded Parameter Norms}

The following table provides an example training log from Experiment 1 (IMDb, $T = 350$). The spectral abscissa $s_{\bm{A}}$ is already plotted in Figure 1 (bottom middle, orange line). At the beginning of a stable training cycle, in which $s_{\bm{A}}$ is successfully pushed toward $0$ from above and the loss drops significantly within 10 epochs, the parameter norms do not show strictly increasing trends. This supports the claim that these norms remain bounded. Additional logs are available in our GitHub repository.

\vspace{8pt}

\begin{table}[h!]
\centering
\resizebox{\textwidth}{!}{
\begin{tabular}{|c|c|c|c|c|c|c|c|c|c|c|}
\hline
\textbf{epoch} & $s_{\bm{A}}$ & $|p|$ & $\|q\|_2$ & $\|W_B\|_2$ & $\|W_B\|_{1,1}$ & $\|W_C\|_2$ & $\|W_C\|_{1,1}$ & $\|A_c\|_2$ & $\max{\|u[t]\|_2}$ & \textbf{Loss} \\ \hline
0  & 0.100  & 1.43  & 4.12  & 7.68  & 51.0  & 7.40  & 48.5  & 9.51  & 7.39  & -              \\ \hline
1  & 0.069  & 1.46  & 3.89  & 7.12  & 45.9  & 6.85  & 43.4  & 9.41  & 7.07  & $6.4 \times 10^{20}$ \\ \hline
3  & 0.005  & 1.62  & 3.56  & 6.53  & 41.4  & 6.33  & 39.1  & 9.21  & 6.49  & 347.7          \\ \hline
5  & -0.006 & 1.58  & 3.44  & 6.21  & 39.5  & 6.15  & 39.0  & 9.01  & 6.19  & 2.04           \\ \hline
7  & -0.009 & 1.58  & 3.37  & 6.05  & 38.6  & 6.13  & 39.0  & 8.84  & 6.00  & 1.00           \\ \hline
9  & -0.010 & 1.61  & 3.33  & 5.91  & 37.5  & 6.12  & 39.0  & 8.66  & 5.82  & 0.67           \\ \hline
\end{tabular}
}
\vspace{8pt}
\caption{\textbf{IMDb}. Parameter logs during training epochs for Experiment 1, $T=350$.}
\end{table}

\subsection{Experiment 1: Sweeping Spectral Abscissa with Constant Sequence Length}

\begin{figure*}[h]
    \centering

    \begin{minipage}[t]{0.31\textwidth}
        \centering
        \includegraphics[width=\linewidth]{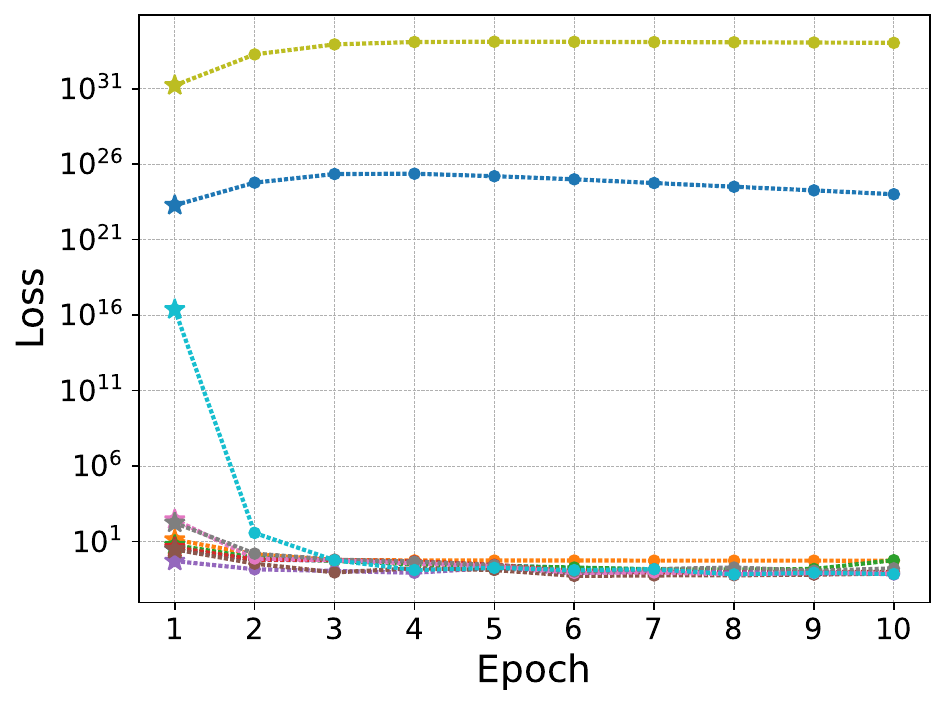} 
    \end{minipage}
    \begin{minipage}[t]{0.31\textwidth}
        \centering
        \includegraphics[width=\linewidth]{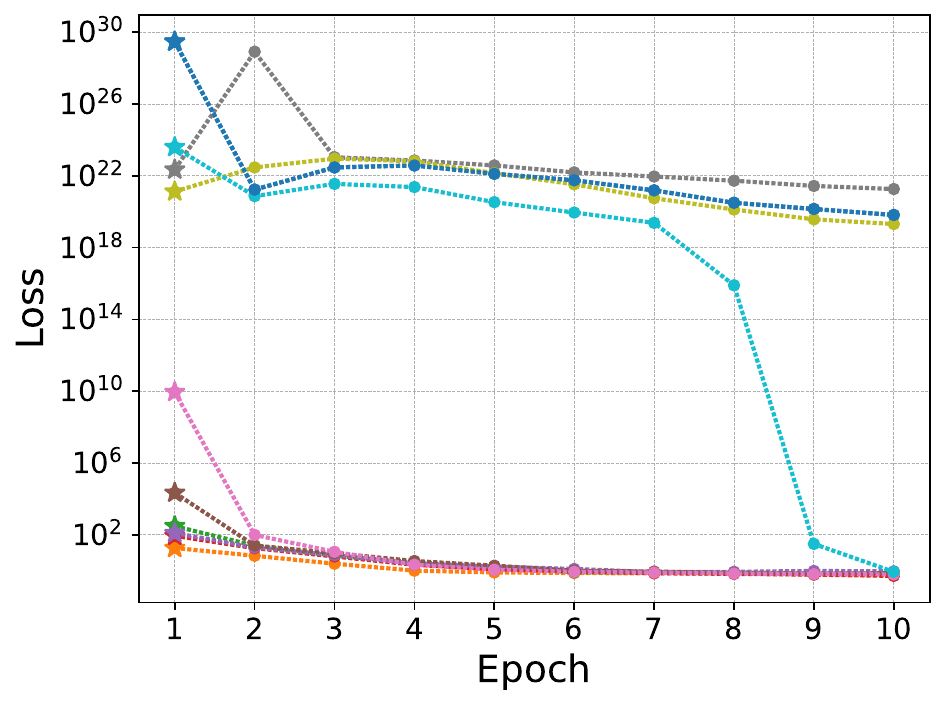} 
    \end{minipage}
    \begin{minipage}[t]{0.31\textwidth}
        \centering
        \includegraphics[width=\linewidth]{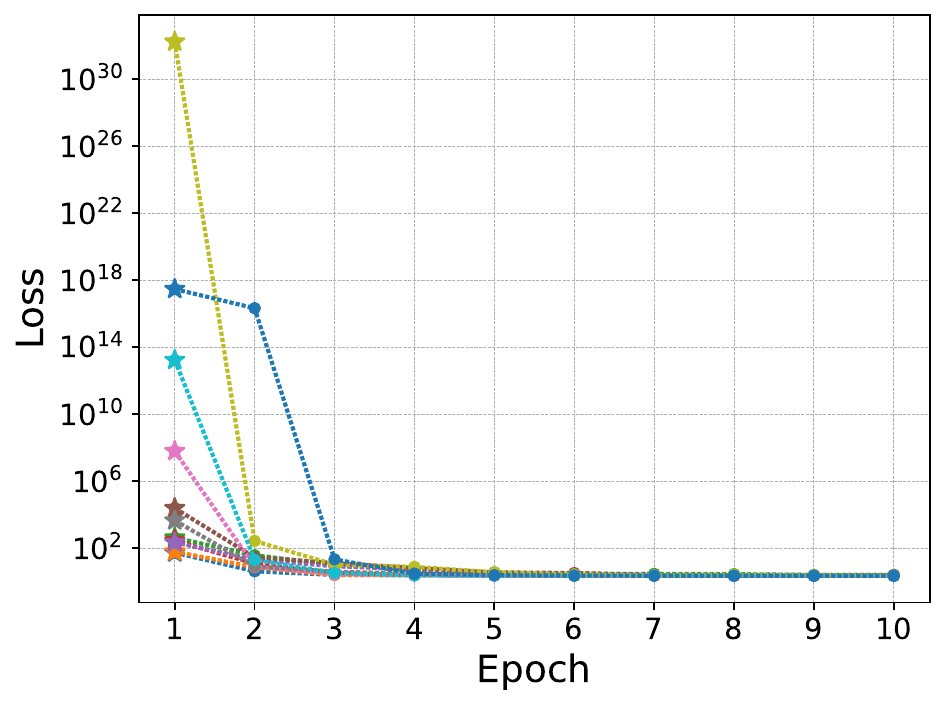} 
    \end{minipage}

    \vspace{0.8em} 

    \begin{minipage}[t]{0.32\textwidth}
        \centering
        \includegraphics[width=\linewidth]{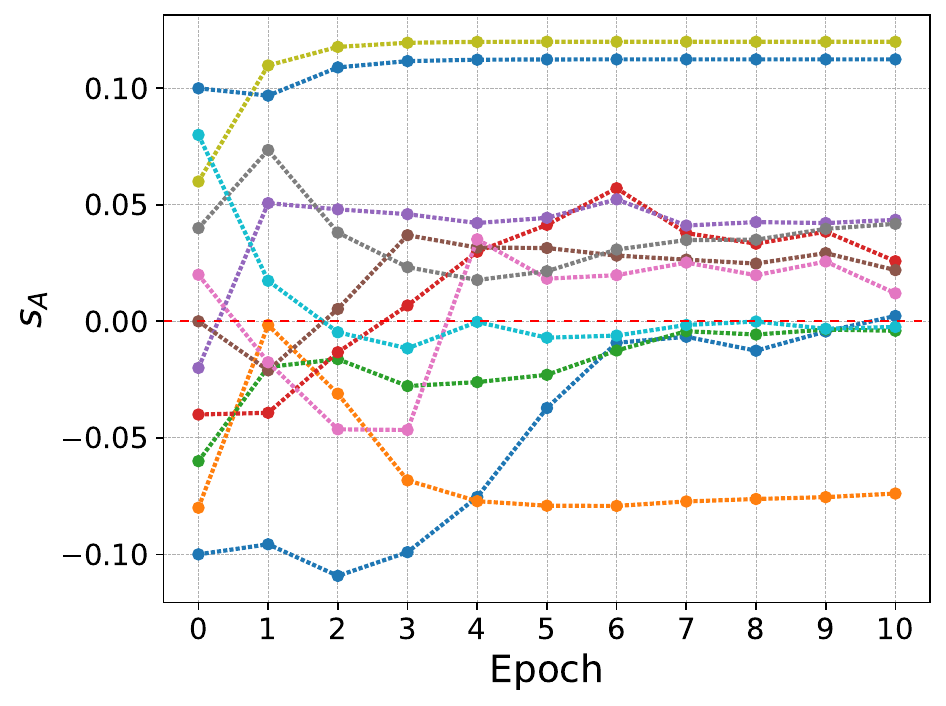} 
    \end{minipage}
    \begin{minipage}[t]{0.32\textwidth}
        \centering
        \includegraphics[width=\linewidth]{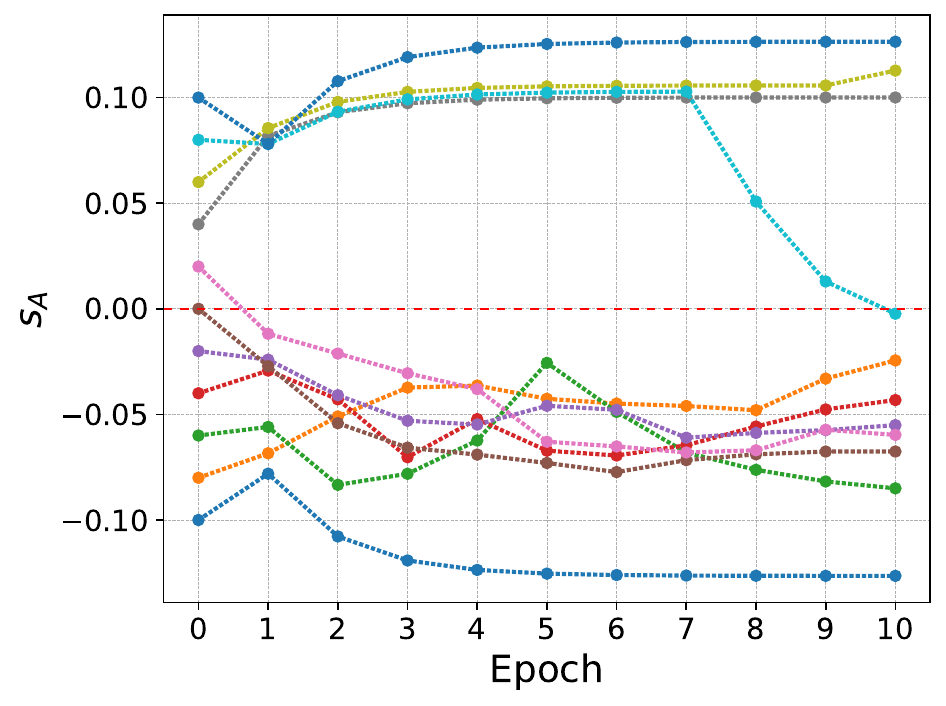} 
    \end{minipage}
    \begin{minipage}[t]{0.32\textwidth}
        \centering
        \includegraphics[width=\linewidth]{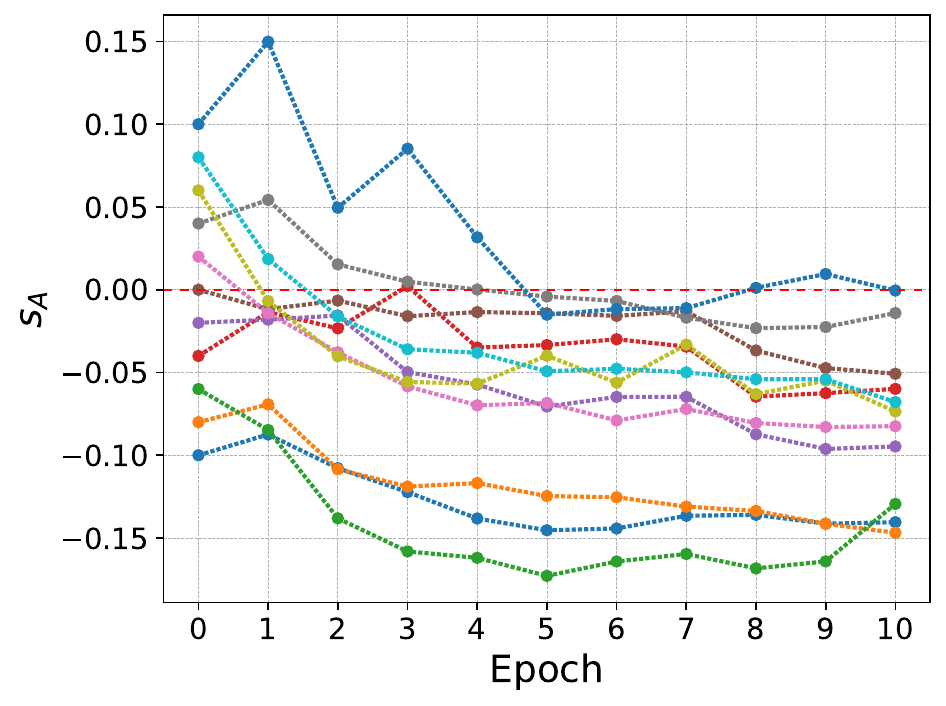} 
    \end{minipage}

    \caption{\small{\textbf{Experiment 1.} \textbf{Top:} Training loss vs epochs for \textbf{Left:} Majority, $T=250$, \textbf{Middle:} IMDb, $T=500$, \textbf{Right:} ListOps, $T=300$.
    \textbf{Bottom:} Evolution of $s_{\bm{A}}$ vs epochs for the same datasets. We sweep the  $s_{\bm{A}}$ values from $-0.1$ to $0.1$ in $0.02$ increments.}}
    \label{fig:sweep_sA}
\end{figure*}

\vspace{8pt}

To understand the effect of different initialization regimes in more detail, we conduct Experiment 1 again, but this time we sweep the $s_{\bm{A}}$ values while keeping $T$ fixed. In order to observe the full effect without encountering runs that diverge (i.e., contain NaN losses), we choose $T=250$ for Majority, $T=500$ for IMDb, and $T=300$ for ListOps. The results in Figure~\ref{fig:sweep_sA} are consistent with those in Figure~\ref{fig:unstable_behavior}. Unstable initializations (where $s_{\bm{A}} > 0$) often lead to training failure due to exploding losses, as other parameters have not yet adapted to stabilize the system. In cases where training succeeds, $s_{\bm{A}}$ is quickly pushed toward zero from above, restoring stability. In contrast, stable initializations ($s_{\bm{A}} < 0$) enable smooth training and remain stable throughout for IMDb and ListOps.

For Majority, stable initializations below zero are also pushed toward zero, and some even end up slightly above it (around $s_{\bm{A}} \approx 0.05$), which is acceptable since other parameters ($\mathfrak{B}_q$, $\mathfrak{B}_u$, and $p$) can compensate according to~\eqref{eq:rho_Mdelta_S}. The key insight is that initialization determines whether training can proceed stably. When $s_{\bm{A}}$ starts in an unstable region, large initial losses arise before other parameters can adapt, often causing divergence. In contrast, initializing in the stable regime enables smooth optimization and coordinated parameter adaptation, allowing $\rho_{\bm{A}}$ to remain balanced through~\eqref{eq:rho_Mdelta_S}. Thus, proper initialization of $s_{\bm{A}}$ is essential for both convergence and stable expressivity.

\section{Useful Lemmas}
\label{sec:useful_lemmas}

\begin{lemma}
\label{lem:B_kronecker_bound}
    Let $\bm{B}_c = \mathbf{I}_d \otimes \bm{W_B} u[t]$. Then, $\| \bm{B}_c \|_2 \leq \mathfrak{B}_{\bm{B}} \mathfrak{B}_u$.
\end{lemma}
\begin{proof}
    \begin{equation}
    \begin{aligned}
        \| \bm{B}_c \|_2^2 &= \| \mathbf{I}_d \otimes \bm{W_B} u[t] \|_2^2 = \| \bm{W_B} u[t] \|_2^2 \leq \| \bm{W_B} \|_2^2 \mathfrak{B}_u^2 \leq \mathfrak{B}_{\bm{B}}^2 \mathfrak{B}_u^2.
    \end{aligned}
    \end{equation}
\end{proof}

\begin{lemma}
\label{lem:C_kronecker_bound}
    Let $\bm{C}_c = \mathbf{I}_d \otimes u[t]^\top \bm{W_C}^\top$. Then, $\| \bm{C}_c \|_2 \leq \mathfrak{B}_{\bm{C}} \mathfrak{B}_u$.
\end{lemma}
\begin{proof}
    \begin{equation}
    \begin{aligned}
        \| \bm{C}_c \|_2^2 &= \| \mathbf{I}_d \otimes u[t]^\top \bm{W_C}^\top \|_2^2 = \| u[t]^\top \bm{W_C}^\top \|_2^2 = \|\bm{W_C} u[t]\|_2^2
        \leq \| \bm{W_C} \|_2^2 \mathfrak{B}_u^2 \leq \mathfrak{B}_{\bm{C}}^2 \mathfrak{B}_u^2.
    \end{aligned}
    \end{equation}
\end{proof}

\begin{lemma}
\label{lem:e^X-e^Y_bound}
Let $\bm{X}$ and $\bm{Y}$ be matrices such that $\| e^{\bm{X}} \|_2 \leq \rho$ and $\| e^{\bm{Y}} \|_2 \leq \rho$. Then, 
\begin{equation}
    \| e^{\bm{X}} - e^{\bm{Y}} \|_2 \leq \rho \| \bm{Y} - \bm{X} \|_2.
\end{equation}
\end{lemma}

\begin{proof}
Using the fundamental theorem of calculus, we express the difference as
\begin{equation}
    e^{\bm{X}} - e^{\bm{Y}} = \int_0^1 e^{t\bm{Y} + (1 - t)\bm{X}} (\bm{Y} - \bm{X}) \, dt.
\end{equation}
Taking the spectral norm on both sides and applying submultiplicativity, we obtain
\begin{equation}
\begin{aligned}
    \| e^{\bm{X}} - e^{\bm{Y}} \|_2 
    &\leq \left\| \int_0^1 e^{t\bm{Y} + (1 - t)\bm{X}} (\bm{Y} - \bm{X}) \, dt \right\|_2 \\
    &\leq \int_0^1 \| e^{t\bm{Y} + (1 - t)\bm{X}} (\bm{Y} - \bm{X}) \|_2 \, dt \\
    &\leq \| \bm{Y} - \bm{X} \|_2 \int_0^1 \| e^{t\bm{Y} + (1 - t)\bm{X}} \|_2 \, dt \\
    &\leq \rho \| \bm{Y} - \bm{X} \|_2.
\end{aligned}
\end{equation}
\end{proof}

\begin{lemma}
\label{lem:delta_cover}
Let $\hat{p}$ be an $\epsilon_p$-cover for $p$ and $\hat{q}$ be an $\epsilon_q$-cover for $q$. Then,
\begin{equation}
    | \Delta[t] - \hat{\Delta}[t] | \leq \epsilon_p + \epsilon_q.
\end{equation}
\end{lemma}

\begin{proof}
The derivative of the soft plus function, $\ln(1 + e^x)$, is $\frac{e^x}{1 + e^x}$, which is bounded by $1$. Thus, $\ln(1 + e^x)$ is $1$-Lipschitz. using this property, we obtain
\begin{equation}
\begin{aligned}
    | \Delta[t] - \hat{\Delta}[t] | &= \big| \ln(1 + e^{p + q^\top u[t]}) - \ln(1 + e^{\hat{p} + \hat{q}^\top u[t]}) \big| \\
    &\leq \big| p - \hat{p} + (q - \hat{q})^\top u[t] \big| \\
    &\leq | p - \hat{p} | + \| (q - \hat{q})^\top u[t] \|_2.
\end{aligned}
\end{equation}
\end{proof}

\begin{lemma}
\label{lem:small_w_cover}
Let $\hat{w}$ be an $\epsilon_w$-cover for $w$. Then,
\begin{equation}
    \big| w^\top y[T] - \hat{w}^\top \hat{y}[T] \big| \leq \mathfrak{B}_w \| y[T] - \hat{y}[T] \|_2 + \epsilon_w.
\end{equation}
\end{lemma}

\begin{proof}
Rewriting the LHS, we obtain
\begin{equation}
    \big| w^\top y[T] - \hat{w}^\top \hat{y}[T] \big| 
    = \big| w^\top ( y[T] - \hat{y}[T] ) + (w - \hat{w})^\top \hat{y}[T] \big|.
\end{equation}
Applying triangle inequality results in
\begin{equation}
\begin{aligned}
    \big| w^\top y[T] - \hat{w}^\top \hat{y}[T] \big|
    &\leq \| w \|_2 \| y[T] - \hat{y}[T] \|_2 + \big| (w - \hat{w})^\top \hat{y}[T] \big| \\
    &\leq \mathfrak{B}_w \| y[T] - \hat{y}[T] \|_2 + \epsilon_w.
\end{aligned}
\end{equation}
\end{proof}

\begin{lemma}
\label{lem:C_cover}
Let $\bm{\hat{C}}_c$ be an $\epsilon_C$-cover for $\bm{C}_c$. Then,
\begin{equation}
    \| \bm{C}_c x[T] - \bm{\hat{C}}_c \hat{x}[T]\|_2 \leq \mathfrak{B}_{\bm{C}} \mathfrak{B}_u \| x[T] - \hat{x}[T] \|_2 + \epsilon_C.
\end{equation}
\end{lemma}
\begin{proof}
The LHS can be bounded as follows:
    \begin{equation}
    \begin{aligned}
         & \leq \| \bm{C}_c \left( x[T] - \hat{x}[T] \right) + (\bm{C}_c - \bm{\hat{C}}_c) \hat{x}[T] \|_2 \\
         & \leq \| \bm{C}_c \|_2 \| x[T] - \hat{x}[T] \|_2 + \| (\bm{C}_c - \bm{\hat{C}}_c) \hat{x}[T] \|_2.
    \end{aligned}
    \end{equation}
    Applying Lemma \eqref{lem:C_kronecker_bound} completes the proof.
\end{proof}

\begin{lemma}
\label{lem:W_b_cover}
Let $\bm{\hat{W}}_B$ be a cover for $\bm{W}_B$. Then, for any $v \in \mathbb{R}^d$ such that $\| v \|_2 = \mathfrak{B}_v$, we obtain
\begin{equation}
    \| (\bm{B}_c - \bm{\hat{B}}_c) v \|_2 \leq \mathfrak{B}_v \| (\bm{W}_B - \bm{\hat{W}}_B) u \|_2.
\end{equation}
\end{lemma}

\begin{proof}
We use the Kronecker product property $(\bm{X} \otimes \bm{Y}) \operatorname{vec}(\bm{V}) = \operatorname{vec} \left( \bm{Y V X}^T \right)$. Take $\bm{X}$ as $\mathbf{I}_d$, $\bm{V}$ as $v^\top$, and $\bm{Y}$ as $(\bm{W}_B - \bm{\hat{W}}_B) u$ to obtain
\begin{equation}
\begin{aligned}
    \| (\bm{B}_c - \bm{\hat{B}}_c) v \|_2 
    &= \| \left( \mathbf{I}_d \otimes (\bm{W}_B - \bm{\hat{W}}_B) u \right) v \|_2 \\
    &= \| \operatorname{vec} \left( (\bm{W}_B - \bm{\hat{W}}_B ) u v^\top \right) \|_2.
\end{aligned}
\end{equation}
From the definition of the Frobenius norm, we obtain
\begin{equation}
\begin{aligned}
    \| (\bm{W}_B - \bm{\hat{W}}_B ) u v^\top \|_F 
    &\leq \| (\bm{W}_B - \bm{\hat{W}}_B ) u \|_F \| v^\top \|_F \\
    &= \| (\bm{W}_B - \bm{\hat{W}}_B ) u \|_2 \| v \|_2 \\
    &\leq \mathfrak{B}_v \| (\bm{W}_B - \bm{\hat{W}}_B ) u \|_2.
\end{aligned}
\end{equation}
\end{proof}

\begin{lemma}
\label{lem:W_c_cover}
Let $\bm{\hat{W}}_C$ be a cover for $\bm{W}_C$. Then, for any $v \in \mathbb{R}^{Nd}$ such that $\| v \|_2 = \mathfrak{B}_v$, we obtain
\begin{equation}
    \| (\bm{C}_c - \bm{\hat{C}}_c) v \|_2 \leq \mathfrak{B}_v \| (\bm{W}_C - \bm{\hat{W}}_C) u \|_2.
\end{equation}
\end{lemma}
\begin{proof}
    Similar to Lemma~\ref{lem:W_b_cover}.
\end{proof}

\begin{lemma}[\citet{edelman2022inductive}, Lemma A.8]
\label{lem:eq_constr_optim}
For $\alpha_i, \beta_i \geq 0$, the solution to the following optimization
\begin{equation}
\begin{gathered}
    \min _{\epsilon_1, \ldots, \epsilon_n} \sum_{i=1}^n \frac{\alpha_i}{\epsilon_i^2} \\
    \text { subject to } \sum_{i=1}^n \beta_i \epsilon_i=\epsilon
\end{gathered}
\end{equation}
is $\frac{\gamma^3}{\epsilon^2}$ and is achieved at $\epsilon_i=\frac{\epsilon}{\gamma}\left(\frac{\alpha_i}{\beta_i}\right)^{1 / 3}$, where $\gamma=\sum_{i=1}^n \alpha_i^{1 / 3} \beta_i^{\frac{2}{3}}$.
\end{lemma}

\section{Covering Numbers}
\label{sec:covering_numbers}
\begin{lemma}[\citet{bartlett2017spectrally}, Lemma 3.2]
\label{lem:matrix_cover}
    Let conjugate exponents $(p, q)$ and $(r, s)$ be given with $p \leq 2$, as well as positive reals $(a, b, \epsilon)$ and positive integer $d_3$. Let matrix $X \in \mathbb{R}^{d_1 \times d_2}$ be given with $\|X\|_{p,p} \leq b$. Then,
    \begin{equation}
    \ln \mathcal{N} \left(\left\{X A: A \in \mathbb{R}^{d_2 \times d_3},\|A\|_{q, s} \leq a \right\}, \epsilon,\|\cdot\|_F \right) \leq \left\lceil\frac{a^2 b^2 d_3^{2 / r}}{\epsilon^2}\right\rceil \ln (2 d_2 d_3).
    \end{equation}
\end{lemma}

\begin{lemma}
\label{lem:cover_A_selective}
    Let $\mathcal{F}_{\bm{A}_c} = \{ \bm{A}_c \in \mathbb{R}^{Nd \times Nd} : \|\bm{A}_c\|_2 \leq \mathfrak{B}_{\bm{A}} \text{ and } \| \bm{A}_c \|_{2,1} \leq \mathfrak{M}_{\bm{A}} \}$. Then,
    \begin{equation}
        \ln \mathcal{N} (\mathcal{F}_{A_c}, \epsilon_A, \|\cdot\|_2) \leq \frac{2\mathfrak{M}_{A}^2 Nd}{\epsilon_A^2} \ln (\sqrt{2}Nd).
    \end{equation}
\end{lemma}
\begin{proof}
    Note that every $\epsilon_A$-covering number for the Frobenius norm is also 
    an $\epsilon_A$-covering number for the spectral norm, as 
    $\| \bm{A} - \bm{\hat{A}} \|_2 \leq \| \bm{A} - \bm{\hat{A}} \|_F \leq \epsilon_A$. Therefore,
    \begin{equation}
    \begin{aligned}
        \ln \mathcal{N} (\mathcal{F}_{A_c}, \epsilon_A, \|\cdot\|_2) &\leq \ln \mathcal{N} (\mathcal{F}_{A_c}, \epsilon_A, \|\cdot\|_F) \\
        &\leq \ln \mathcal{N} (\{ \bm{A}_c \in \mathbb{R}^{Nd \times Nd} : \| \bm{A}_c \|_{2,1} \leq \mathfrak{M}_{\bm{A}} \}, \epsilon_A, \|\cdot\|_F).
    \end{aligned}
    \end{equation}
    Thus, we instantiate Lemma \ref{lem:matrix_cover} with $p=q=2$ and $s=1, r=\infty$. Take $X$ to be identity and thus $b = \sqrt{Nd}$ which results in
    \begin{equation}
        \ln \mathcal{N} (\{ \bm{A}_c \in \mathbb{R}^{Nd \times Nd} : \| \bm{A}_c \|_{2,1} \leq \mathfrak{M}_{\bm{A}} \}, \epsilon_A, \|\cdot\|_F) \leq  \left\lceil\frac{\mathfrak{M}_{A}^2 (\sqrt{Nd})^2}{\epsilon_A^2}\right\rceil \ln (2 Nd Nd).
    \end{equation}
\end{proof}

\begin{lemma}[\citet{trauger2024length_independent_transformer}, Lemma 3.6]
\label{lem:linear_func_cover}
    Let $m \geq d_2$, $\mathcal{F}_W = \{ \bm{W}u : \bm{W} \in \mathbb{R}^{d_1 \times d_2}, \| \bm{W} \|_{1,1} \leq \mathfrak{M}_W \}$. If $\| u \|_2 \leq \mathfrak{B}_u$, then
    \begin{equation}
        \ln \mathcal{N}_\infty (\mathcal{F}_W, \epsilon_W, \|\cdot\|_2) \leq \frac{\mathfrak{B}_u^2 \mathfrak{M}_W^2}{\epsilon_W^2} \ln(2 d_1 d_2 + 1).
    \end{equation}
\end{lemma}
\textbf{Remark.}  
The removal of the dependency on $m$ in the log covering number for a function class is nontrivial and requires specific assumptions about the norm bounds. For similar log covering bounds that are independent of $m$, refer to \cite{trauger2024length_independent_transformer} and the lemmas therein.

\begin{lemma}
\label{lem:generalization_error_bound_last}
    Let $\mathcal{F}$ be a function class such that $\ln \mathcal{N}_\infty (\mathcal{F}, \epsilon, \|\cdot\|_2) \leq \frac{\mathcal{C}_\mathcal{F}^2}{\epsilon^2}$ and let $S$ be the training set $\{u_{(i)}, z_{(i)}\}_{i=1}^m$. Assume the loss function $l:\mathcal{Z} \times \mathcal{Z} \rightarrow \mathbb{R}$ is upper bounded by the constant $\mathfrak{c}_l$ and Lipschitz continuous with constant $\mathfrak{l}_l$. Then, with probability at least $1 - \delta$,
    \begin{equation}
    \begin{aligned}
        \left| \mathbb{E}_{u,z}(l(h(u), z)) - \frac{1}{m} \sum_{i=1}^m l\left(h(u_{(i)}), z_{(i)}\right) \right| \leq \frac{12 \mathfrak{l}_l \mathcal{C}_\mathcal{F}}{\sqrt{m}} \left( 1 + \ln\left(\frac{\mathfrak{c}_l\sqrt{m}}{3 \mathcal{C}_\mathcal{F}}\right) \right) + 3\mathfrak{c}_l \sqrt{\frac{\ln\left(\frac{2}{\delta}\right)}{2m}}.
    \end{aligned}
    \end{equation}
\end{lemma}
\begin{proof}
    By Theorem~\ref{thm:dudley}, and the fact that $\ln \mathcal{N}_2 (l \circ \mathcal{F}, \epsilon, \|\cdot\|_2) \leq \ln \mathcal{N}_\infty (l \circ \mathcal{F}, \epsilon, \|\cdot\|_2)$ (check Definition 1 in \cite{zhang2002covering}), we have
    \begin{equation}
        \operatorname{Rad}(l \circ \mathcal{F}, S) \leq \inf_{\alpha > 0} \left( 4 \alpha + 12 \int_\alpha^{\mathfrak{c}_l} \sqrt{\frac{\ln \mathcal{N}_\infty (l \circ \mathcal{F}, \epsilon, \|\cdot\|_2)}{m}} \; d\epsilon \right).
    \end{equation}
    Upper bound $\ln \mathcal{N}_\infty (\mathcal{F}, \epsilon, m, \| \cdot \|_2)$ as specified by the lemma to obtain
    \begin{equation}
    \label{eq:inf_sth_over_sqrt_m}
    \begin{aligned}
        &\leq \inf_{\alpha > 0} \left( 4\alpha + \frac{12}{\sqrt{m}} \int_\alpha^{\mathfrak{c}_l} \frac{\mathfrak{l}_l \mathcal{C}_\mathcal{F}}{\epsilon} \; d\epsilon \right) =  \inf_{\alpha > 0} \left( 4 \alpha + \frac{12 \mathfrak{l}_l \mathcal{C}_\mathcal{F}}{\sqrt{m}} \ln\left(\frac{\mathfrak{c}_l}{\alpha}\right) \right)
    \end{aligned}
    \end{equation}
    in which we used $\ln \mathcal{N}_\infty (l \circ \mathcal{F}, \epsilon, \|\cdot\|_2) \leq \mathfrak{l}_l \ln \mathcal{N}_\infty (\mathcal{F}, \epsilon, \|\cdot\|_2)$. The minimum of \eqref{eq:inf_sth_over_sqrt_m} occurs at $\alpha = \frac{3 \mathfrak{l}_l \mathcal{C}_\mathcal{F}}{\sqrt{m}}$. Thus,
    \begin{equation}
        \leq \frac{12 \mathfrak{l}_l \mathcal{C}_\mathcal{F}}{\sqrt{m}} + \frac{12 \mathfrak{l}_l \mathcal{C}_\mathcal{F}}{\sqrt{m}} \ln\left(\frac{\mathfrak{c}_l\sqrt{m}}{3 \mathcal{C}_\mathcal{F}}\right) = \frac{12 \mathfrak{l}_l \mathcal{C}_\mathcal{F}}{\sqrt{m}} \left( 1 + \ln\left(\frac{\mathfrak{c}_l\sqrt{m}}{3 \mathcal{C}_\mathcal{F}}\right) \right).
    \end{equation}
    Combining this bound on the Rademacher complexity with Theorem~\ref{thm:gen_error_rademacher} concludes the proof.
\end{proof}

\section{\texorpdfstring{Proof for Theorem~\ref{thm:gen_err_bound_selective}: Generalization Error Bound for Selective SSMs}{Proof for Theorem: Generalization Error Bound for Selective SSMs}}
\label{sec:input_dependent_gen_err_proof}
Before presenting the proof of Theorem~\ref{thm:gen_err_bound_selective}, we provide preliminary background, including the definition of Rademacher complexity and a standard theorem establishing its connection to the generalization gap. Then, we introduce four intermediate lemmas tailored to the selective SSMs. The first lemma establishes an upper bound on the spectral norm of the state  matrix after $t$ repetitions, namely $\| \bm{A}^t \|_2$. The second lemma bounds the distance between the time-varying product $\bm{A}^t$ and its corresponding cover, accounting for the input-dependent nature of the state matrices. These two lemmas are necessary and specific to selective SSMs, since, unlike standard RNNs, the state matrix $\bm{A}$ is not fixed. As a result, classical norm bounds do not directly apply, and we must instead derive upper bounds in terms of the model parameters $\bm{A}_c$, $p$, and $q$, which govern the input-dependent dynamics. The third and fourth lemmas build upon the first two to inductively bound the distance between the output of a selective SSM and that of its corresponding cover. These results culminate in the proof of Theorem~\ref{thm:gen_err_bound_selective} provided at the end of this section.

\begin{definition}[\textbf{Rademacher complexity}]
    For a given real-valued Function class $\mathcal{F}$ and a set of vectors $S = \{u_{(i)}\}_{i=1}^m$, the empirical Rademacher complexity is
\begin{equation}
    \operatorname{Rad}(\mathcal{F}, S) = \frac{1}{m} \mathbb{E}_\sigma \left( \sup_{f \in \mathcal{F}} \sum_{i=1}^m \sigma_i f(u_{(i)}) \right)
    \label{eq:rademacher}
\end{equation}
where $\sigma_i\in\{-1,1\}$ are uniformly distributed i.i.d Rademacher random variables.
\label{def:rademacher}
\end{definition}

\begin{theorem}[\citet{mohri2018foundations_machine_learning}, Theorem 3.3]
\label{thm:gen_error_rademacher}
Let $\mathcal{F}$ be a hypothsis class $\{ f:\mathcal{U
}\rightarrow\mathcal{Z}\}$, and $S$ be the training set $\{u_{(i)}, z_{(i)}\}_{i=1}^m$. Assume the loss function $l:\mathcal{Z} \times \mathcal{Z} \rightarrow \mathbb{R}$ is upper bounded by the constant $\mathfrak{c}_l$. Then, with probability more than $1 - \delta$
    \begin{equation}
    \begin{aligned}
        &\left| \mathbb{E}_{u,z}(l(f(u), z)) - \frac{1}{m} \sum_{i=1}^m l\left(f(u_{(i)}), z_{(i)} \right) \right| \leq 2 \operatorname{Rad}(l \circ \mathcal{F}, S) + 3\mathfrak{c}_l \sqrt{\frac{\ln(\frac{2}{\delta})}{2m}}.
    \end{aligned}
    \end{equation}
\end{theorem}

\begin{lemma}
\label{lem:A^t_rho^t_bound_selective}
Let $s_{\bm{A}}$ be the spectral abscissa of  $\bm{A}_c$ i.e. $\max_i \Re(\lambda_i(\bm{A}_c))$. Suppose $u[t] \leq \mathfrak{B}_u$ and $\| q \|_2 \leq \mathfrak{B}_q$. Then, given any arbitrary small $\eta> 0$, there exists a sufficiently large $t$ such that
\begin{equation}
    \| \bm{A}^t \|_2 \leq \rho_{\bm{A}}^t,
\end{equation}
where
\begin{equation}
    \rho_{\bm{A}} = \left( 1 + e^{p - \mathfrak{B}_q \mathfrak{B}_u} \right)^{s_{\bm{A}} + \eta}.
\end{equation}
\end{lemma}

\begin{proof}
    From \eqref{eq:A^t_shorthand}, we have
    \begin{equation}
        \| \bm{A}^t \|_2 = \left\| \prod_{j=T-t}^{T-1} e^{\Delta[j] \bm{A}_c} \right\|_2.
    \end{equation}
    Given the assumptions of the lemma, and noting that the softplus function, 
    $\ln(1 + e^x)$, is increasing, we derive the following lower bound:
    \begin{equation}
        \Delta[j] \geq \ln(1 + e^{p - \mathfrak{B}_q \mathfrak{B}_u}).
    \end{equation}
    Since the spectral abscissa of $\bm{A}_c$ is $s_{\bm{A}}$, the spectral radius of $e^{\bm{A}_c}$ would be $e^{s_{\bm{A}}}$. By Gelfand's formula (Corollary 5.6.14 in \cite{horn2012matrix}), we have that $\| (e^{\bm{A}_c})^t \|_2^{1/t} \to e^{s_{\bm{A}}}$ as $t \to \infty$. Consequently, for an arbitrary small positive number $\eta > 0$, there exists a sufficiently large $t_0$ such that for all $t \geq t_0$, the following bound holds:
    \[
    \| (e^{\bm{A}_c})^t \|_2 \leq \left( e^{s_{\bm{A}} + \eta} \right)^t.
    \]
    This yields the desired exponential norm bound for large $t$:
    \begin{equation}
    \begin{aligned}
        \| \bm{A}^t \|_2 &= \left\| \left(e^{\bm{A}_c}\right)^{\sum_{j=T-t}^{T-1} \Delta[j]} \right\|_2 \\
        &\leq \left(e^{s_{\bm{A}}+\eta}\right)^{t \ln\left(1 + e^{p - \mathfrak{B}_q \mathfrak{B}_u}\right)} \\
        &= \left( 1 + e^{p - \mathfrak{B}_q \mathfrak{B}_u} \right)^{(s_{\bm{A}}+\eta) t} = \rho_{\bm{A}}^t.
    \end{aligned}
    \end{equation}
\end{proof}

\begin{lemma}
\label{lem:upper_bound_x[T]}
    We have the following upper bound on $x[T]$:
    \begin{equation}
        \| x[T] \|_2 \leq \mathfrak{M}_\Delta \mathfrak{B}_{\bm{B}} \mathfrak{B}_u^2 \frac{1 - \rho_A^T}{1 - \rho_A}
    \end{equation}
    in which $\rho_{\bm{A}} = \left( 1 + e^{p - \mathfrak{B}_q \mathfrak{B}_u} \right)^{s_A + \eta}$ and $\mathfrak{M}_\Delta = \ln(1 + e^{p + \mathfrak{B}_q \mathfrak{B}_u})$.
\end{lemma}
\begin{proof}
    \begin{equation}
    \begin{aligned}
        \| x[T] \|_2 &\leq \left\| \sum_{t=0}^{T-1} \Big( \bm{A}^t \Delta[T - 1 - t] \big( \mathbf{I}_d \otimes \bm{W_B} u[T-1-t] \big) u[T-1-t] \Big) \right\|_2 \\
        & \leq  \sum_{t=0}^{T-1} \Big( \| \bm{A}^t \|_2 \| \Delta[T - 1 - t] \big( \mathbf{I}_d \otimes \bm{W_B} u[T-1-t] \big) \|_2 \| u[T-1-t] \|_2 \Big) \\
        & \leq  \mathfrak{M}_\Delta \mathfrak{B}_u \sum_{t=0}^{T-1} \Big( \| \bm{A}^t \|_2 \| \mathbf{I}_d \otimes \bm{W_B} u[T-1-t] \|_2 \Big) \\
        & \leq \mathfrak{M}_\Delta \mathfrak{B}_{\bm{B}} \mathfrak{B}_u^2 \sum_{t=0}^{T-1} \rho_{\bm{A}}^t = \mathfrak{M}_\Delta \mathfrak{B}_{\bm{B}} \mathfrak{B}_u^2 \frac{1 - \rho_A^T}{1 - \rho_A},
    \end{aligned}
    \end{equation}
    where we used Lemma~\ref{lem:A^t_rho^t_bound_selective} to bound $\| \bm{A}^t \|_2$ and Lemma~\ref{lem:B_kronecker_bound} to upper bound the term involving the Kronecker product to derive the last inequality.
\end{proof}

\begin{lemma}
\label{lem:A^t-Ahat^t_bound_selective}
Let $\bm{\hat{A}}_c$ be an $\epsilon_A$-cover for $\bm{A}_c$, $\hat{p}$ be an $\epsilon_p$-cover for $p$, and $\hat{q}$ be an $\epsilon_q$-cover for $q$. Then,
\begin{equation}
    \| \bm{A}^t - \bm{\hat{A}}^t \|_2 \leq t \rho_{\bm{A}}^t (\mathfrak{M}_{\Delta} \epsilon_A + \mathfrak{B}_{\bm{A}} \epsilon_\Delta )
\end{equation}
where $\mathfrak{M}_{\Delta} = \ln(1 + e^{p + \mathfrak{B}_q \mathfrak{B}_u})$.
\end{lemma}

\begin{proof}
We start with
\begin{equation}
    \begin{aligned}
        \| \bm{A}^t - \bm{\hat{A}}^t \|_2 &= \left\| \prod_{k=T-t}^{T-1} e^{\Delta[k] \bm{A}_c}-\prod_{k=T-t}^{T-1} e^{\hat{\Delta}[k] \bm{\hat{A}}_c} \right\|_2 \\
        &= \left\| \sum_{i=T-t}^{T-1} \left( \left(\prod_{j=i}^{T-1} e^{\Delta[j] \bm{A}_c}\right) \left(\prod_{k=T-t}^{i-1} e^{\hat{\Delta}[k] \bm{\hat{A}}_c}\right) \right. \right. \\
        & \qquad \qquad \left. \left. -  \left(\prod_{j=i+1}^{T-1} e^{\Delta[j] \bm{A}_c}\right) \left(\prod_{k=T-t}^{i} e^{\hat{\Delta}[k] \bm{\hat{A}}_c}\right) \right) \right\|_2.
    \end{aligned}
\end{equation}
Factor common terms to obtain
\begin{equation}
    \begin{aligned}
        &\leq \left\| \sum_{i=T-t}^{T-1} \left(\prod_{j=i+1}^{T-1} e^{\Delta[j] \bm{A}_c}\right) \left(e^{\Delta[i] \bm{A}_c}-e^{\hat{\Delta}[i] \bm{\hat{A}}_c}\right) \left(\prod_{k=T-t}^{i-1} e^{\hat{\Delta}[k] \bm{\hat{A}}_c}\right) \right\|_2 \\
        &\leq \sum_{i=T-t}^{T-1} \left\| \prod_{j=i+1}^{T-1} e^{\Delta[j] \bm{A}_c} \right\|_2 \left\| e^{\Delta[i] \bm{A}_c} - e^{\hat{\Delta}[i] \bm{\hat{A}}_c} \right\|_2 \left\| \prod_{k=T-t}^{i-1} e^{\hat{\Delta}[k] \bm{\hat{A}}_c} \right\|_2.
    \end{aligned}
\end{equation}
Applying Lemma \ref{lem:A^t_rho^t_bound_selective}, we get 
\begin{equation}
    \begin{aligned}
        &\leq \sum_{i=T-t}^{T-1} \rho_{\bm{A}}^{T-i-1} \| e^{\Delta[i] \bm{A}_c} - e^{\hat{\Delta}[i] \bm{\hat{A}}_c} \|_2 \rho_{\bm{A}}^{i-T+t} \\
        &= \sum_{i=T-t}^{T-1} \rho_{\bm{A}}^{t-1} \| e^{\Delta[i] \bm{A}_c} - e^{\hat{\Delta}[i] \bm{\hat{A}}_c} \|_2.
    \end{aligned}
\end{equation}
Use Lemma \ref{lem:e^X-e^Y_bound} to derive
\begin{equation}
    \| \bm{A}^t - \bm{\hat{A}}^t \|_2 \leq \rho_{\bm{A}}^t \sum_{i=T-t}^{T-1} \| \Delta[i] \bm{A}_c - \hat{\Delta}[i] \bm{\hat{A}}_c \|_2.
\end{equation}
Apply triangle inequality:
\begin{equation}
    \begin{aligned}
        &\leq \rho_{\bm{A}}^t \sum_{i=T-t}^{T-1} \left( \| \Delta[i] (\bm{A}_c - \bm{\hat{A}}_c) \|_2 + \| (\Delta[i] - \hat{\Delta}[i]) \bm{\hat{A}}_c \|_2 \right) \\
        &\leq \rho_{\bm{A}}^t \sum_{i=T-t}^{T-1} \left( \mathfrak{M}_{\Delta} \epsilon_A + | \Delta[i] - \hat{\Delta}[i] | \mathfrak{B}_{\bm{A}} \right) \\
        &\leq \rho_{\bm{A}}^t \sum_{i=T-t}^{T-1} \left( \mathfrak{M}_{\Delta} \epsilon_A + \epsilon_\Delta \mathfrak{B}_{\bm{A}} \right).
    \end{aligned}
\end{equation}
At last, we obtain the final bound:
\begin{equation}
    \| \bm{A}^t - \bm{\hat{A}}^t \|_2 \leq t \rho_{\bm{A}}^t (\mathfrak{M}_{\Delta} \epsilon_A + \mathfrak{B}_{\bm{A}} \epsilon_\Delta).
\end{equation}
\end{proof}

\begin{lemma}
\label{lem:A_cover_selective}
Let $\bm{\hat{A}}, \bm{\hat{B}}, \hat{\Delta}$ be covers for $\bm{A}, \bm{B}, \Delta$. Then,
\begin{equation}
\begin{aligned}
    & \left\| \sum_{t=0}^{T-1} \bm{A}^t \Delta[T-1-t] \bm{B}_c u[T-1-t] - \sum_{t=0}^{T-1} \bm{\hat{A}}^t \hat{\Delta}[T-1-t] \bm{\hat{B}}_c u[T-1-t] \right\|_2 \\
    & \leq \mathfrak{M}_\Delta S_1 \epsilon_B + \mathfrak{M}_\Delta^2 \mathfrak{B}_{\bm{B}} \mathfrak{B}_u^2 S_2 \epsilon_A
    + \mathfrak{B}_{\bm{B}} \mathfrak{B}_u^2 \left( S_1 + \mathfrak{M}_\Delta \mathfrak{B}_{\bm{A}} S_2 \right) \epsilon_{\Delta}
\end{aligned}
\end{equation}
, where $S_1 = \frac{1 - \rho_{\bm{A}}^T}{1 - \rho_{\bm{A}}}$ and $S_2 = \frac{\rho_{\bm{A}}(1 - \rho_{\bm{A}}^T)}{(1 - \rho_{\bm{A}})^2} - \frac{T \rho_{\bm{A}}^T}{1 - \rho_{\bm{A}}}$.
\end{lemma}

\begin{proof}
    Write the LHS as follows:
    \begin{equation}
        \left\| \sum_{t=0}^{T-1} \left( \left( \bm{A}^t \Delta[T-1-t] \bm{B}_c - \bm{\hat{A}}^t \hat{\Delta}[T-1-t] \bm{\hat{B}}_c \right) u[T-1-t] \right) \right\|_2.
    \end{equation}
    Add and subtract the terms $\sum_{t=0}^{T-1}  \left( \bm{A}^t \Delta[T-1-t] \bm{\hat{B}}_c \right) u[t-t-1]$ and $\sum_{t=0}^{T-1}  \left( \bm{A}^t \hat{\Delta}[T-1-t] \bm{\hat{B}}_c \right) u[t-t-1]$ to derive
    \begin{equation}
    \begin{aligned}
        \| &\sum_{t=0}^{T-1} \left( \bm{A}^t \Delta[T-1-t] \bm{B}_c - \bm{A}^t \Delta[T-1-t] \bm{\hat{B}}_c \right) u[T-t-1] \\
          +&\sum_{t=0}^{T-1} \left( \bm{A}^t \Delta[T-1-t] \bm{\hat{B}}_c - \bm{A}^t \hat{\Delta}[T-1-t] \bm{\hat{B}}_c \right) u[T-t-1] \\
          +&\sum_{t=0}^{T-1} \left( \bm{A}^t \hat{\Delta}[T-1-t] \bm{\hat{B}}_c - \bm{\hat{A}}^t \hat{\Delta}[T-1-t] \bm{\hat{B}}_c \right) u[T-t-1] \|_2.
    \end{aligned}
    \end{equation}
    Apply the triangle inequality to get
    \begin{equation}
    \begin{aligned}
        \sum_{t=0}^{T-1} \Big( & \| \bm{A}^t \Delta[T-1-t] ( \bm{B}_c - \bm{\hat{B}}_c ) u[T-t-1] \|_2 \Big. \\ 
                         \Big. & + \| \bm{A}^t (\Delta[T-1-t] - \hat{\Delta}[T-1-t]) \bm{\hat{B}}_c u[T-t-1] \|_2 \Big. \\
                         \Big. & + \| ( \bm{A}^t - \bm{\hat{A}}^t) \hat{\Delta[T-1-t]} \bm{\hat{B}}_c u[T-t-1] \|_2 \Big)
    \end{aligned}
    \end{equation}
    which is upper bounded by
    \begin{equation}
    \begin{aligned}
        \leq \sum_{t=0}^{T-1} \Big( & \| \bm{A}^t \|_2 | \Delta[T-1-t] | \| ( \bm{B}_c - \bm{\hat{B}}_c ) u[T-t-1] \|_2 \Big. \\ 
                         \Big. & + \| \bm{A}^t \|_2 | \Delta[T-1-t] - \hat{\Delta}[T-1-t] | \| \bm{\hat{B}}_c u[T-t-1] \|_2 \Big. \\
                         \Big. & + \| ( \bm{A}^t - \bm{\hat{A}}^t) \|_2 | \hat{\Delta}[T-1-t] |\| \bm{\hat{B}}_c u[T-t-1] \|_2 \Big).
    \end{aligned}
    \end{equation}
    The application of Lemmas \ref{lem:A^t_rho^t_bound_selective} and \ref{lem:A^t-Ahat^t_bound_selective} to bound $\|\bm{A}^t\|_2$ and cover $\| \bm{A}^t - \bm{\hat{A}}^t \|_2$ results in
    \begin{equation}
    \begin{aligned}
        \leq \sum_{t=0}^{T-1} \Big( \rho_{\bm{A}}^t \mathfrak{M}_\Delta \epsilon_B \Big.
                         \Big. + \rho_{\bm{A}}^t \epsilon_\Delta \| \bm{\hat{B}}_c \|_2 \mathfrak{B}_u \Big.
                         \Big. + t \rho_{\bm{A}}^t (\mathfrak{M}_{\Delta} \epsilon_A + \mathfrak{B}_{\bm{A}} \epsilon_\Delta) \mathfrak{M}_\Delta \| \bm{\hat{B}}_c \|_2 \mathfrak{B}_u \Big).
    \end{aligned}
    \end{equation}
    Apply Lemma \ref{lem:B_kronecker_bound} to bound $\| \bm{\hat{B}}_c \|_2$:
    \begin{equation}
        \leq \sum_{t=0}^{T-1} \left( \rho_{\bm{A}}^t \mathfrak{M}_\Delta \epsilon_B + \rho_{\bm{A}}^t \mathfrak{B}_{\bm{B}} \mathfrak{B}_u^2 \epsilon_\Delta + t \rho_{\bm{A}}^t \mathfrak{M}_\Delta\mathfrak{B}_{\bm{B}} \mathfrak{B}_u^2 (\mathfrak{M}_{\Delta} \epsilon_A + \mathfrak{B}_{\bm{A}} \epsilon_\Delta) \right).
    \end{equation}
    Breaking the summation into two parts leads to
    \begin{equation*}
    \begin{aligned}
        &\leq \left( \mathfrak{M}_\Delta \epsilon_B + \mathfrak{B}_{\bm{B}} \mathfrak{B}_u^2 \epsilon_\Delta  \right) \sum_{t=0}^{T-1} \rho_{\bm{A}}^t + \mathfrak{M}_\Delta \mathfrak{B}_{\bm{B}} \mathfrak{B}_u^2 (\mathfrak{M}_{\Delta} \epsilon_A + \mathfrak{B}_{\bm{A}} \epsilon_\Delta) \sum_{t=0}^{T-1} t \rho_{\bm{A}}^t \\
        &\leq \left( \mathfrak{M}_\Delta \epsilon_B + \mathfrak{B}_{\bm{B}} \mathfrak{B}_u^2 \epsilon_\Delta \right) \frac{1 - \rho_{\bm{A}}^T}{1 - \rho_{\bm{A}}} + \mathfrak{M}_\Delta \mathfrak{B}_{\bm{B}} \mathfrak{B}_u^2 (\mathfrak{M}_{\Delta} \epsilon_A + \mathfrak{B}_{\bm{A}} \epsilon_\Delta) \left( \frac{\rho_{\bm{A}} (1 - \rho_{\bm{A}}^T)}{(1 - \rho_{\bm{A}})^2} - \frac{T \rho_{\bm{A}}^T}{1 - \rho_{\bm{A}}} \right)
    \end{aligned}
    \end{equation*}
    which completes the proof.
\end{proof}

\textbf{Remark.}\;
Lemma~\ref{lem:A^t_rho^t_bound_selective} is stated for sufficiently
large $t$. In Lemmas \ref{lem:A^t-Ahat^t_bound_selective} and \ref{lem:A_cover_selective} we still apply that bound when summing over all time indices. The step is legitimate because any sum $\sum_{t=0}^{T-1}(\cdot)$ can be split at an index $t_0$ for which the
hypothesis of Lemma~\ref{lem:A^t_rho^t_bound_selective} holds for $t \geq t_0$:
\[
\sum_{t=0}^{T-1}(\cdot) = \sum_{t=0}^{t_0-1}(\cdot) + \sum_{t=t_0}^{T-1}(\cdot).
\]
The first term involves only finitely many values of $t$ and therefore
contributes a constant that is absorbed into the leading
$\mathcal O(\,\cdot\,)$ rate. The second (tail) term is where the bound of
Lemma~\ref{lem:A^t_rho^t_bound_selective} is used, and it determines the asymptotic dependence on~$T$. Hence, omitting the constant part does not affect the final
Big-$\mathcal O$ expression in the main theorem.

\begin{lemma}
\label{lem:epsilon_sum_bound_selective}
Let $\bm{\hat{A}}_c, \bm{\hat{W}_B}, \bm{\hat{W}_C}, \hat{p}, \hat{q}$ be covers for $\bm{A}_c, \bm{W_B}, \bm{W_C}, p, q$. Then,
\begin{equation}
\label{eq:eps_cover_decompose_params}
\begin{aligned}
    \left| w^\top y[T] - \hat{w}^\top \hat{y}[T] \right| 
    &\leq \mathfrak{B}_w \mathfrak{B}_{\bm{C}} \mathfrak{B}_u^2 \mathfrak{M}_\Delta S_1 \epsilon_{W_B} 
    + \mathfrak{M}_\Delta^2 \mathfrak{B}_w \mathfrak{B}_{\bm{B}} \mathfrak{B}_{\bm{C}} \mathfrak{B}_u^3 S_2 \epsilon_A \\
    &\quad + \left( \mathfrak{B}_w \mathfrak{B}_{\bm{B}} \mathfrak{B}_{\bm{C}} \mathfrak{B}_u^3 \right) 
     \left( S_1 + \mathfrak{M}_\Delta \mathfrak{B}_{\bm{A}} S_2 \right) (\epsilon_p + \epsilon_q) \\
    &\quad + \mathfrak{B}_w \mathfrak{B}_u \epsilon_{W_C} + \epsilon_w,
\end{aligned}
\end{equation}
where $S_1$ and $S_2$ are defined as in Lemma~\ref{lem:A_cover_selective}.
\end{lemma}

\begin{proof}
The proof follows from the sequential application of Lemmas \ref{lem:small_w_cover}, \ref{lem:C_cover}, and \ref{lem:A_cover_selective}, yielding:
\begin{equation}
\begin{aligned}
    &\left| w^\top y[T] - \hat{w}^\top \hat{y}[T] \right| \\
    &\leq \mathfrak{B}_w \left( \mathfrak{B}_{\bm{C}} \mathfrak{B}_u 
    \left( \mathfrak{M}_\Delta S_1 \epsilon_B + \mathfrak{M}_\Delta^2 \mathfrak{B}_{\bm{B}} \mathfrak{B}_u^2 S_2 \epsilon_A
    + \mathfrak{B}_{\bm{B}} \mathfrak{B}_u^2 \left( S_1 + \mathfrak{M}_\Delta \mathfrak{B}_{\bm{A}} S_2 \right) \epsilon_{\Delta} \right) 
    + \epsilon_C \right) + \epsilon_w.
\end{aligned}
\end{equation}
Finally, we apply Lemmas \ref{lem:W_b_cover}, \ref{lem:W_c_cover}, and \ref{lem:delta_cover} to relate the covers for $\bm{B}, \bm{C}, \bm{\Delta}$ to the covers for $\bm{W_B}, \bm{W_C}, p, q$, completing the proof.
\end{proof}

\begin{proof}[Proof of Theorem \ref{thm:gen_err_bound_selective}]
    We aim to construct a cover for the space of all selective SSMs $\mathcal{F}_{\text{SSM}} = \{ z[T] = w^\top y[T] : y[T] \text{ is described in \eqref{eq:unrolled_io_selective}} \}$ which is parametrizes by $\Theta_{\text{SSM}} = \{\bm{A}_c, \bm{W_B}, \bm{W_C}, p, q, w\}$. Let's look at how much the output $w^\top y[T]$ changes if we move to the points in the $\epsilon$-net constructing the cover. This is done in Lemma \ref{lem:epsilon_sum_bound_selective}. Thus, we need to choose $\epsilon_A, \epsilon_{W_B}, \epsilon_{w_C}$, $\epsilon_q$, $\epsilon_p$ and $\epsilon_w$ subject to the following:
    \begin{equation}
    \begin{aligned}
        \epsilon &= \mathfrak{B}_w \mathfrak{B}_{\bm{C}} \mathfrak{B}_u^2 \mathfrak{M}_\Delta S_1 \epsilon_{W_B} 
        + \mathfrak{M}_\Delta^2 \mathfrak{B}_w \mathfrak{B}_{\bm{B}} \mathfrak{B}_{\bm{C}} \mathfrak{B}_u^3 S_2 \epsilon_A \\
        &\quad + \left( \mathfrak{B}_w \mathfrak{B}_{\bm{B}} \mathfrak{B}_{\bm{C}} \mathfrak{B}_u^3 \right) 
         \left( S_1 + \mathfrak{M}_\Delta \mathfrak{B}_{\bm{A}} S_2 \right) (\epsilon_p + \epsilon_q) \\
        &\quad + \mathfrak{B}_w \mathfrak{B}_u \epsilon_{W_C} + \epsilon_w,
    \end{aligned}
    \end{equation}
    which relates the $\epsilon$-cover of a selective SSM to corresponding covers for each parameter in $\Theta_{\text{SSM}}$ as in \eqref{eq:eps_cover_decompose_params}.
    
    Choose the covering for $\bm{W}_B$ according to Lemma \ref{lem:linear_func_cover} such that
    \begin{equation}
    \begin{aligned}
        \ln \mathcal{N}_\infty (\mathcal{F}_{W_B}, \epsilon_{W_B}, \|\cdot\|_2) \leq \frac{\mathfrak{B}_u^2 \mathfrak{M}_{\bm{B}}^2}{\epsilon_{W_B}} \ln(2Nd + 1).
    \end{aligned}
    \end{equation}
    Similarly, choose the covering for $\bm{W}_C$ by replacing $v$ in Lemma~\ref{lem:linear_func_cover} with $x[T]$ which is bounded as in Lemma~\ref{lem:upper_bound_x[T]} to derive
    \begin{equation}
    \begin{aligned}
        \ln \mathcal{N}_\infty (\mathcal{F}_{W_C}, \epsilon_{W_C}, \|\cdot\|_2) &\leq \frac{\left( \mathfrak{M}_\Delta \mathfrak{B}_{\bm{B}} \mathfrak{B}_u^2 S_1 \right)^2 \mathfrak{M}_{\bm{C}}^2}{\epsilon_{W_C}^2} \ln(2dNd + 1) \\
        &= \frac{\mathfrak{B}_{\bm{B}}^2 \mathfrak{B}_u^4 \mathfrak{M}_\Delta^2 \mathfrak{M}_{\bm{C}}^2 S_1^2}{\epsilon_{_{W_C}}^2} \ln(2Nd^2 + 1).
    \end{aligned}
    \end{equation}
    Likewise, choose the cover for $w$ such that
    \begin{equation}
    \begin{aligned}
        &\ln \mathcal{N}_\infty (\mathcal{F}_w, \epsilon_w, \|\cdot\|_2) \leq \frac{\left(  \mathfrak{M}_\Delta \mathfrak{B}_{\bm{C}} \mathfrak{B}_{\bm{B}} \mathfrak{B}_u^3 S_1  \right)^2 \mathfrak{M}_w^2}{\epsilon_w^2} \ln(2d + 1) \\
        &= \frac{ \mathfrak{B}_{\bm{B}}^2 \mathfrak{B}_{\bm{C}}^2 \mathfrak{B}_u^6 \mathfrak{M}_\Delta^2 \mathfrak{M}_w^2 S_1^2}{ \epsilon_w^2 } \ln(2d + 1).
    \end{aligned}
    \end{equation}
    Lemma~\ref{lem:cover_A_selective} gives us the upper bound on the covering number for $\bm{A}_c$:
    \begin{equation}
        \ln \mathcal{N} (\mathcal{F}_{A_c}, \epsilon_A, \|\cdot\|_2) \leq \frac{2\mathfrak{M}_{A}^2 Nd}{\epsilon_A^2} \ln (\sqrt{2}Nd).
    \end{equation}
    We may use Lemma~\ref{lem:linear_func_cover} again to cover $q$:
    \begin{equation}
        \ln \mathcal{N}_\infty (\mathcal{F}_{q}, \epsilon_q, \|\cdot\|_2) \leq \frac{\mathfrak{B}_u^2 \mathfrak{M}_q^2}{\epsilon_q^2} \ln(2d+1)
    \end{equation}
    and $p$ is covered simply by
    \begin{equation}
        \mathcal{N}_\infty (\mathcal{F}_{p}, \epsilon_p, \|\cdot\|_2) \leq \frac{2|p|}{\epsilon_p}.
    \end{equation}
    Ignore the logarithmic dependencies and assume $\mathfrak{M}_{\bm{C}} = \mathfrak{B}_{\bm{C}}, \mathfrak{M}_{\bm{B}} = \mathfrak{B}_{\bm{B}}, \mathfrak{M}_w = \mathfrak{B}_w, \mathfrak{M}_q = \mathfrak{B}_q, \mathfrak{M}_{\bm{A}} = \mathfrak{B}_{\bm{A}}$ for simplicity. Construct the cover for the space of all selective SSMs $\mathcal{F}_{\text{SSM}}$ as the Cartesian product of all covers for each parameter in $\Theta_{\text{SSM}}$. Then, the log covering number would be the sum of the log covering numbers of all parameters. Use Lemma \ref{lem:eq_constr_optim} to find $\epsilon_A, \epsilon_{W_B}, \epsilon_{w_C}, \epsilon_q, \epsilon_q$, and $\epsilon_w$ such that the size of total cover would be minimum:
    \begin{equation}
    \label{eq:SSM_cov_num_first_upper_bound}
    \begin{aligned}
        & \epsilon^2 \ln \mathcal{N}_\infty (\mathcal{F}_{\text{SSM}}, \epsilon, \|\cdot\|_2) \\
        &\leq \mathcal{\tilde{O}} \left( \left( (\mathfrak{B}_u^2 \mathfrak{B}_{\bm{B}}^2)^{1/3} (\mathfrak{B}_w \mathfrak{B}_{\bm{C}} \mathfrak{B}_u^2 \mathfrak{M}_\Delta S_1)^{2/3} + ( \mathfrak{B}_{\bm{B}}^2 \mathfrak{B}_u^4 \mathfrak{M}_\Delta^2 \mathfrak{B}_{\bm{C}}^2 S_1^2 )^{1/3}(\mathfrak{B}_w \mathfrak{B}_u)^{2/3} \right. \right. \\
        & + (\mathfrak{B}_{\bm{A}}^2 N d)^{1/3} (\mathfrak{M}_\Delta^2 \mathfrak{B}_w \mathfrak{B}_{\bm{B}} \mathfrak{B}_{\bm{C}} \mathfrak{B}_u^3 S_2)^{2/3} + (\mathfrak{B}_{\bm{B}}^2 \mathfrak{B}_{\bm{C}}^2 \mathfrak{B}_u^6 \mathfrak{M}_\Delta^2 \mathfrak{B}_w^2 S_1^2 )^{1/3} \\
        & + \left. \left. (\mathfrak{B}_u^2 \mathfrak{B}_q^2)^{1/3} \left( \mathfrak{B}_w \mathfrak{B}_{\bm{B}} \mathfrak{B}_{\bm{C}} \mathfrak{B}_u^3 \right)^{2/3}
          \left( S_1 + \mathfrak{M}_\Delta \mathfrak{B}_{\bm{A}} S_2 \right)^{2/3} \right)^3 \right).
    \end{aligned}
    \end{equation}
    in which we ignored the cover for $p$ as it is dominated by other terms.
    \begin{equation}
    \begin{aligned}
        &\leq \mathcal{\tilde{O}} \left( \left( \mathfrak{M}_\Delta^{2/3} \mathfrak{B}_w^{2/3} \mathfrak{B}_u^2 \mathfrak{B}_{\bm{B}}^{2/3} \mathfrak{B}_{\bm{C}}^{2/3} S_1^{2/3} + \mathfrak{M}_\Delta^{2/3} \mathfrak{B}_w^{2/3} \mathfrak{B}_u^2 \mathfrak{B}_{\bm{B}}^{2/3} \mathfrak{B}_{\bm{C}}^{2/3} S_1^{2/3} \right. \right. \\
        & + \mathfrak{M}_\Delta^{4/3} \mathfrak{B}_{\bm{A}}^{2/3} \mathfrak{B}_w^{2/3} \mathfrak{B}_u^2 \mathfrak{B}_{\bm{B}}^{2/3} \mathfrak{B}_{\bm{C}}^{2/3} S_2^{2/3} N^{1/3} d^{1/3} + \mathfrak{M}_\Delta^{2/3} \mathfrak{B}_w^{2/3} \mathfrak{B}_u^2 \mathfrak{B}_{\bm{B}}^{2/3} \mathfrak{B}_{\bm{C}}^{2/3} S_1^{2/3} \\
        & + \left. \left. \mathfrak{M}_\Delta^{2/3} \mathfrak{B}_{\bm{A}}^{2/3} \mathfrak{B}_q^{2/3} \mathfrak{B}_w^{2/3} \mathfrak{B}_u^{8/3} \mathfrak{B}_{\bm{B}}^{2/3} \mathfrak{B}_{\bm{C}}^{2/3} S_2^{2/3} \right)^3 \right),
    \end{aligned}
    \end{equation}
    where we used the fact that $S_1$ is dominated by $S_2$ for large $T$ to obtain the last term. Therefore, we have
    \begin{equation}
    \begin{aligned}
        &\leq \mathcal{\tilde{O}} \left( \mathfrak{M}_\Delta^2 \mathfrak{B}_w^{2} \mathfrak{B}_u^6 \mathfrak{B}_{\bm{B}}^{2} \mathfrak{B}_{\bm{C}}^{2} \left( S_1^{2/3} + S_1^{2/3} + \mathfrak{M}_\Delta^{2/3} \mathfrak{B}_{\bm{A}}^{2/3} S_2^{2/3} N^{1/3} d^{1/3} \right. \right. \\
        &\qquad \left. \left. + 1 + \mathfrak{B}_{\bm{A}}^{2/3} \mathfrak{B}_q^{2/3} \mathfrak{B}_u^{2/3} S_2^{2/3} \right)^3 \right).
    \end{aligned}
    \end{equation}
    Ignoring the constant terms and $S_1$ compared to $S_2$ results in 
    \begin{equation}
        \leq \mathcal{\tilde{O}} \left( \mathfrak{M}_\Delta^2 \mathfrak{B}_w^{2} \mathfrak{B}_u^6 \mathfrak{B}_{\bm{B}}^{2} \mathfrak{B}_{\bm{C}}^{2} \mathfrak{B}_{\bm{A}}^2 S_2^2 (\mathfrak{M}_\Delta^{2/3} N^{1/3} d^{1/3} + \mathfrak{B}_q^{2/3} \mathfrak{B}_u^{2/3} )^3 \right).
    \end{equation}
    The square root of this expression is $\mathcal{C}_\mathcal{F}$. The proof is complete by the application of Lemma \ref{lem:generalization_error_bound_last}.
\end{proof}

\section{\texorpdfstring{Proof for Proposition~\ref{prop:gen_error_linear_attention}: Genralization Error Bound for Linear Attentions }{Proof for Propostion: Genralization Error Bound for Linear Attentions}}
\label{sec:lin_att_gen_err}

\begin{proof}[Proof of Proposition~\ref{prop:gen_error_linear_attention}]
    The proof that follows is similar to the proof of Theorem~\ref{thm:gen_err_bound_selective} with modifications accounting for the simplifications made in Assumptions \ref{ass:constant_step_size} and \ref{ass:identity_state_matrix}.  Specifically, we do not need to cover $\bm{A}_c$, $p$, or $q$.  Therefore, Lemma~\ref{lem:A_cover_selective} simplifies to
    \begin{equation}
        \Big\|\sum_{t=0}^{T-1} (\bm{W}_B - \widehat{\bm{W}}_B)\,u[T-1-t]\Big\|_2
        \;\le\;
        T\,\epsilon_{W_B}.
    \end{equation}
    Consequently, Lemma~\ref{lem:epsilon_sum_bound_selective} becomes
    \begin{equation}
        \bigl|w^\top y[T] - \widehat w^\top \widehat y[T]\bigr|
        \;\le\;
        T\,\mathfrak{B}_w\,\mathfrak{B}_{\bm{C}}\,\mathfrak{B}_u^2\,\epsilon_{W_B}
        \;+\;
        \mathfrak{B}_w\,\mathfrak{B}_u\,\epsilon_{W_C}
        \;+\;
        \epsilon_w.
    \end{equation}
    Also, Lemma~\ref{lem:upper_bound_x[T]} reduces to
    \begin{equation}
        \|x[T]\|_2 \;\le\; T\,\mathfrak{B}_{\bm{B}}\,\mathfrak{B}_u^2,
    \end{equation}
    which is used to cover $\bm W_C$.  Hence \eqref{eq:SSM_cov_num_first_upper_bound} becomes
    \begin{equation}
    \begin{aligned}
        \epsilon^2 \ln \mathcal{N}_\infty(\mathcal{F}_{LA}, \epsilon, \|\cdot\|_2)
        &\le
        \widetilde{\mathcal{O}}\Bigl(\,
          (\mathfrak{B}_u^2 \mathfrak{B}_{\bm{B}}^2)^{\tfrac13}
          (T\,\mathfrak{B}_w \mathfrak{B}_{\bm{C}} \mathfrak{B}_u^2)^{\tfrac23}\\
        &\quad
          +\,
          (T^2\,\mathfrak{B}_{\bm{B}}^2 \mathfrak{B}_u^4 \mathfrak{B}_{\bm{C}}^2)^{\tfrac13}
          (\mathfrak{B}_w \mathfrak{B}_u)^{\tfrac23}
          +\,
          (T^2\,\mathfrak{B}_{\bm{B}}^2 \mathfrak{B}_{\bm{C}}^2 \mathfrak{B}_u^6 \mathfrak{B}_w^2)^{\tfrac13}
        \Bigr)^3 \\
        &=
        \widetilde{\mathcal{O}}\bigl(T^2\,\mathfrak{B}_{\bm{B}}^2\,\mathfrak{B}_{\bm{C}}^2\,\mathfrak{B}_u^6\,\mathfrak{B}_w^2\bigr).
    \end{aligned}
    \end{equation}
    By applying Lemma~\ref{lem:generalization_error_bound_last}, the proof is complete.
\end{proof}

\section{\texorpdfstring{Proof for Theorem~\ref{thm:lower_bound}: Lower Bound on the Rademacher Complexity}{Proof for Theorem: Lower Bound on the Rademacher Complexity}}
\label{app:lower_bound}

\begin{proof}[Proof of Theorem \ref{thm:lower_bound}]
We consider a restricted class of scalar ($d=N=1$) selective SSMs defined by
\begin{equation}
\Theta_l = \left\{ \bm{A}_c = \ln(1 + s_{\bm{A}}),\;
\bm{W}_B = 1,\;
\bm{W}_C = 1,\;
p = e,\;
q = 0,\;
w \; \bigm| \; |w| \le \mathfrak{B}_w \right\}.
\end{equation}
Since $\Theta_l \subset \Theta_{\text{SSM}}$, the Rademacher complexity of this restricted class provides a lower bound for that of the full class $\mathcal{F}_{\text{SSM}}$. For this class the step size would be fixed $\Delta[k] = 1$, and the discretized matrices become
\begin{equation}
\bm{A}[k] = 1 + s_{\bm{A}}, \quad
\bm{B}[k] = u[k], \quad
\bm{C}[k] = u[k].
\end{equation}
Thus, the resulting state space recurrence is
\begin{equation}
x[k] = (1 + s_{\bm{A}}) x[k-1] + u[k]^2, \qquad
y[k] = u[k] x[k], \qquad
z[k] = w y[k].
\end{equation}
With constant input $u[k] = 1$, the closed-form expression for the output becomes
\begin{equation}
z[T] = w \frac{(1 + s_{\bm{A}})^T - 1}{s_{\bm{A}}}.
\end{equation}
Hence, the hypothesis class can be expressed as
\begin{equation}
\mathcal{F}_l = \left\{ w \frac{(1 + s_{\bm{A}})^T - 1}{s_{\bm{A}}} \; \bigm| \; |w| \le \mathfrak{B}_w \right\}.
\end{equation}
The empirical Rademacher complexity is then
\begin{equation}
\operatorname{Rad}_{\mathcal{S}}(\mathcal{F}_l)
= \frac{1}{m} \mathbb{E}_{\sigma}
\left[ \sup_{|w| \le \mathfrak{B}_w} \sum_{i=1}^{m} \sigma_i z_{(i)}[T] \right]
= \frac{(1 + s_{\bm{A}})^T - 1}{s_{\bm{A}}} \frac{1}{m} \mathbb{E}_{\sigma}
\left[ \sup_{|w| \le \mathfrak{B}_w} w \sum_{i=1}^{m} \sigma_i \right].
\end{equation}
The supremum is achieved when $w = \mathfrak{B}_w \operatorname{sign}\left(\sum_{i=1}^m \sigma_i\right)$, yielding
\begin{equation}
\operatorname{Rad}_{\mathcal{S}}(\mathcal{F}_l) = \mathfrak{B}_w \frac{(1 + s_{\bm{A}})^T - 1}{s_{\bm{A}}} \sqrt{\frac{2}{\pi m}}.
\end{equation}

When $s_{\bm{A}} = 0$, the recursion becomes $x[k] = x[k-1] + 1$, so $x[T] = T$, and hence $z[T] = w T$. The resulting hypothesis class is $\{ w T \mid |w| \le \mathfrak{B}_w \}$, and a similar argument yields
\begin{equation}
\operatorname{Rad}_{\mathcal{S}}(\mathcal{F}_l) = \mathfrak{B}_w T \sqrt{\frac{2}{\pi m}}.
\end{equation}
\end{proof}

\section{Extra Discussion}

Note that the term $S_2$ in Theorem \ref{thm:gen_err_bound_selective}, derived in Lemma D.6) is
\[
S_2
= \frac{\rho_A\!\left(1-\rho_A^{T}\right)}{(1-\rho_A)^2}
  - \frac{T\,\rho_A^{T}}{1-\rho_A}.
\]
This expression does not diverge as $\rho_A \to 1$ even though there is a $1-\rho_A$ in the denominator. Applying L'H\^opital's rule twice yields
\[
S_2=\frac{T^2-T}{2},
\]
which is finite.
Therefore, although increasing $\rho_A$ degrades generalization, the generalization bound does not blow up at $\rho_A=1$; rather, it remains bounded and grows quadratically with $T$. The generalization becomes severely damaged only when $\rho_A>1$, where the bound becomes exponentially increasing in $\mathcal{O}(T \rho_A^{T})$.

\paragraph{Proof that the bound remains finite as $\rho \to 1$.}
We compute
\begin{align*}
\lim_{\rho\to 1}\!\left(
\frac{\rho(1-\rho^{T})}{(1-\rho)^2}
-\frac{T\rho^{T}}{1-\rho}
\right)
&=
\lim_{\rho\to 1}
\frac{\rho - T\rho^{T} + (T-1)\rho^{T+1}}{(1-\rho)^2}.
\end{align*}
First derivatives:
\[
\frac{d}{d\rho}\bigl(\text{numerator}\bigr)
= 1 - T^{2}\rho^{T-1} + (T^{2}-1)\rho^{T},
\qquad
\frac{d}{d\rho}\bigl(\text{denominator}\bigr)
= -2(1-\rho).
\]
Second derivatives:
\[
\frac{d^{2}}{d\rho^{2}}\bigl(\text{numerator}\bigr)
= -T^{2}(T-1)\rho^{T-2} + T(T^{2}-1)\rho^{T-1},
\qquad
\frac{d^{2}}{d\rho^{2}}\bigl(\text{denominator}\bigr)
= 2.
\]
Evaluating at $\rho\to 1$ gives
\[
\lim_{\rho\to 1}
\left(
\frac{\rho(1-\rho^{T})}{(1-\rho)^2}
-\frac{T\rho^{T}}{1-\rho}
\right)
= \frac{T^{2}-T}{2}.
\]

\end{document}